\documentclass{article}

\usepackage{arxiv}

\usepackage[utf8]{inputenc} 
\usepackage[T1]{fontenc}    
\usepackage{hyperref}       
\usepackage{url}            
\usepackage{booktabs}       
\usepackage{amsfonts}       
\usepackage{amsmath,amssymb}                
\usepackage{microtype}      
\usepackage{cleveref}       
\usepackage{graphicx}
\usepackage{subcaption}
\usepackage{doi}
\usepackage[dvipsnames]{xcolor}             
\usepackage[numbers,sort&compress]{natbib}  
\usepackage{enumitem}
\usepackage{arydshln}
\usepackage{multirow}

\title{Attitude-Aided Linear Calibration of Triaxial Accelerometers}

\author{
    Yongqiang Yu$^{1,*}$
    \And
    Tian Huang$^{1}$
    \And
    Yipeng Yang$^{2}$
}
\renewcommand{\corauthor}{
    $^{1}$Independent researcher\\
    $^{2}$Harbin Institute of Technology\\
    $^{*}$Corresponding author: \texttt{winkin119@gmail.com}
}

\renewcommand{\shorttitle}{ALAC: Attitude-Aided Linear Accelerometer Calibration}

\hypersetup{
    pdftitle={Attitude-Aided Linear Calibration of Triaxial Accelerometers},
    pdfsubject={eess.SP, cs.RO},
    pdfauthor={Yongqiang Yu, Tian Huang},
    pdfkeywords={
        ALAC, Closed-form, sensor fusion, least-squares estimation, SVD, Non-iterative,
        Online calibration, MEMS IMU, Robotics, Turntable
    },
    colorlinks=true,
    linkcolor=Red,
    citecolor=Green,
    urlcolor=Blue
}

\begin{document}
\maketitle

\begin{abstract}
Triaxial microelectromechanical systems (MEMS) accelerometers are widely utilized for inertial
sensing, navigation, and sensor fusion. However, existing calibration methods either rely
on costly reference setups or require nonlinear iterative optimization, limiting their efficiency
and general applicability, particularly for low-cost or self-calibrating sensor systems.
We present an attitude-aided, \emph{linear} accelerometers calibration (ALAC) method
that operates on any platform providing orientation information,
such as a turntable, a robotic arm, or an inertial measurement unit (IMU).
The proposed method constructs a combined error matrix (CEM) to represent sensor errors in a
unified calibration model, enabling linear least-squares estimation. The bias and gravity vector are
jointly estimated, implicitly accounting for platform misalignment. A subsequent matrix
decomposition of the CEM yields the scale, non-orthogonality, and alignment rotation parameters.
Calibration is formulated as a constrained homogeneous least-squares (CHLS) problem under static gravity,
and solved in closed-form, non-iterative fashion using standard linear algebra.
A minimum of five arbitrarily oriented measurements suffices, and a recursive
extension enables online or in-field calibration.
Experiments on a stationary robot-mounted accelerometer and a quasi-static IMU trajectory from a
publicly available dataset show that ALAC, in both offline and online modes, outperforms baseline reference-based and online
calibration methods in both accuracy and robustness to sensor noise.
On the same dataset, it further yields accuracy on par with iterative self-calibration methods
under filtered conditions and surpasses all evaluated baselines on raw measurements.
The results demonstrate robustness to system uncertainties,
providing a practical calibration scheme for MEMS-based inertial platforms such as traditional
turntable and low-cost IMUs, particularly suited for online calibration.
The source code is available at \href{https://github.com/GeneHit/ALAC}{https://github.com/GeneHit/ALAC}.
\end{abstract}


\section{Introduction}
With the rapid development of MEMS, MEMS accelerometers, typically
integrated with gyroscopes and magnetometers into IMUs, have been
widely applied in aerospace, robotics, medicine, portable electronic devices, and other fields
\cite{Ahmad2013IMUReview, Johnston2019, zhao2020_satellite, mi12111373, khaula2025_robot}.
These sensors enable accurate motion sensing and positioning in various dynamic environments.
However, their performance is affected by fabrication defects, temperature variations, and
structural limitations, which introduce deterministic and stochastic errors such as offset, drift,
nonlinearity, and random noise~\cite{ru2022mems_survey}. These errors degrade measurement accuracy,
making calibration essential. In particular, low-cost IMUs used in practical applications require
more effective in-field calibration methods~\cite{harindranath2024_survey}.

Accelerometer calibration relies on the static gravity constraint. Depending on whether external
information is employed, existing methods can be categorized as reference-based or self-calibration
approaches.

Self-calibration methods rely solely on accelerometer measurements. These approaches include
least square methods~\cite{rohac2015_newton, lv2016_nl_tls_cuboid, qureshi2017_newton, li2019_newton, sipos2012, tedaldi2014_lm},
maximum likelihood estimation~\cite{frosio2012_lm}, Bayesian estimation~\cite{durr2023_bayesian}
and, more recently, intelligent optimization~\cite{poddar2019_pso, pesti2023_pso} and
neural-network-based methods~\cite{soriano2020, pesti2025_ga_robot}. These methods eliminate the need
for laboratory-grade instruments and can be performed directly in the application environment.
However, they suffer from initialization sensitivity, convergence issues, and local minima due to
model nonlinearity~\cite{ru2022mems_survey, poddar2016_survey}.
Numerous strategies have been developed to mitigate these drawbacks. Multi-position methods, for example,
utilize motion systems such as rotation platforms~\cite{sipos2012} or robot arms~\cite{botero2017lowcost, pesti2025_ga_robot}
to gather high-quality data~\cite{cai2013}.
To improve convergence, a six-tilt-angle initialization was introduced in~\cite{won2010_valid_init},
while scale factors and biases were linearized for computational efficiency in~\cite{ye2017_linearization}.
By contrast, least-squares-based ellipsoid fitting is a linear solution;
however, it cannot estimate inter-trial rotations~\cite{gietzelt201362_ellipsoid}.

Reference-based methods employ positioning systems as external reference sources.
Widely used systems include high-precision turntables or rotation tables
\cite{fang2014_turntable, wang2023mems_turntable}, where sensors are positioned at
predefined orientations and then scale factors and biases are derived through arithmetic and linear
formulations~\cite{cina2019_gnss_meams}.
These methods are primarily applied to high-precision factory or laboratory calibration, where
accuracy relies on precise alignment between the sensor axes and the reference frame~\cite{poddar2016_survey}.
Robotic arms enable rapid and precise sensor repositioning with programmable kinematic control, but
rely on nonlinear calibration models~\cite{Renk2005, beravs2012}.
To lower equipment cost, 3D-printed or custom cuboid fixtures have been adopted as orientation
references, supporting parameter estimation via iterative
optimization~\cite{lv2016_nl_tls_cuboid} or via arithmetic operations~\cite{dong2020_cuboid, xu2021novel}.
Recently, a total least squares (TLS) method was used to estimate partial parameters in~\cite{duchi2023_tls_cuboid}.
An alternative strategy exploits sensor redundancy,
such as using two oppositely mounted accelerometers~\cite{belkhouche2019_2accel} or an external
fiber-optic gyroscope (FOG) IMU~\cite{lu2022_fog_imu}, but they impose accuracy installation.

To overcome the limitations of reference-based methods, we propose an attitude-aided, linear
accelerometer calibration (ALAC) algorithm
for triaxial accelerometers using external attitude measurements. The ALAC is applicable to
any system where attitude information is available, such as a turntable, a robotic arm, or an IMU.
The method adopts a unified calibration model that accounts for biases, scale factors,
non-orthogonality, sensor alignment rotations matrix, and reference platform installation errors.
It requires a minimum of five arbitrarily oriented measurement poses, avoiding the predefined
postures and iterative solvers typical of conventional methods.

Previous studies have demonstrated that online calibration effectively mitigates
time-dependent degradation~\cite{kim2004initial, Renk2005}.
Compared with the complex nonlinear IMU calibration methods~\cite{WANG2017111, ZHANG2019454, troni2019_field, al2023_imu_hand},
our linear method can be readily extended to a recursive online implementation,
enabling dynamic drift compensation that is well known in MEMS devices~\cite{ru2022mems_survey}.
It is computationally efficient and remains free of nonlinear optimization.
\begin{figure}[t]
    \centering
    \begin{minipage}[b]{0.285\linewidth}
        \centering
        \includegraphics[width=\linewidth]{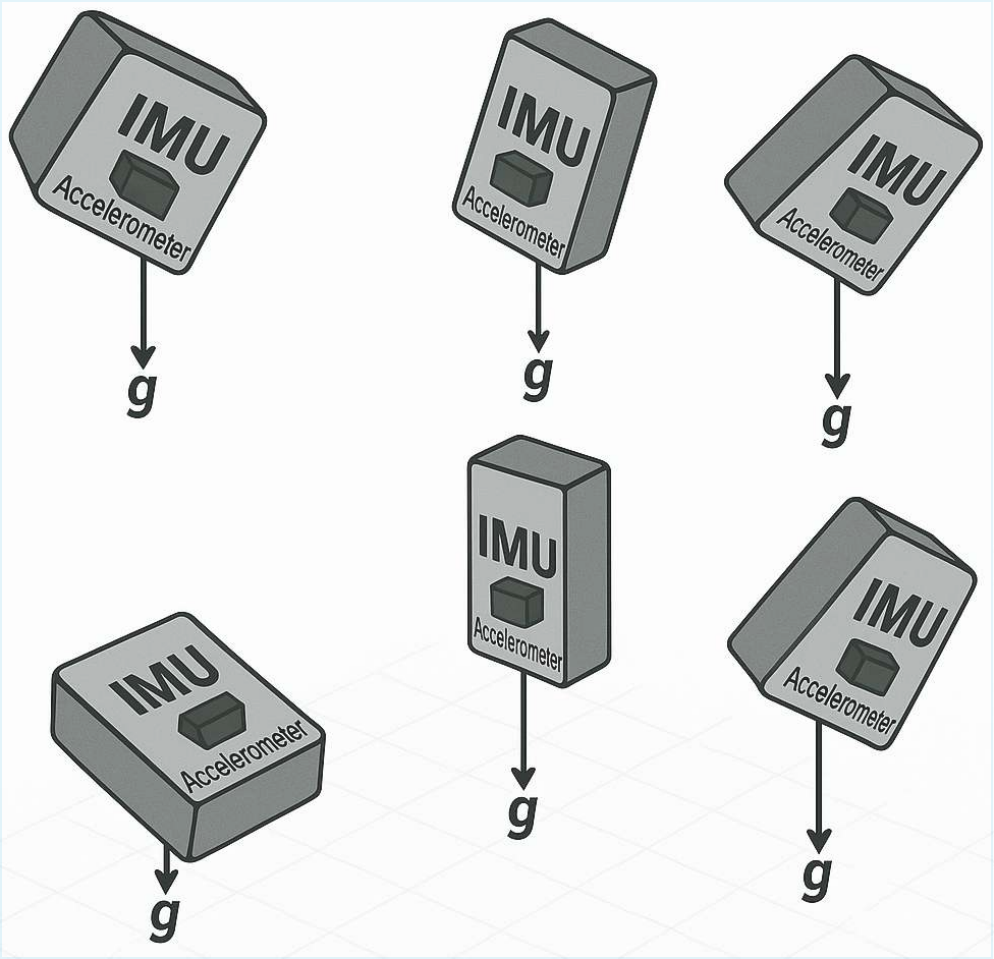}
    \end{minipage}
    \qquad \qquad \qquad
    \begin{minipage}[b]{0.45\linewidth}
        \centering
        \includegraphics[width=\linewidth]{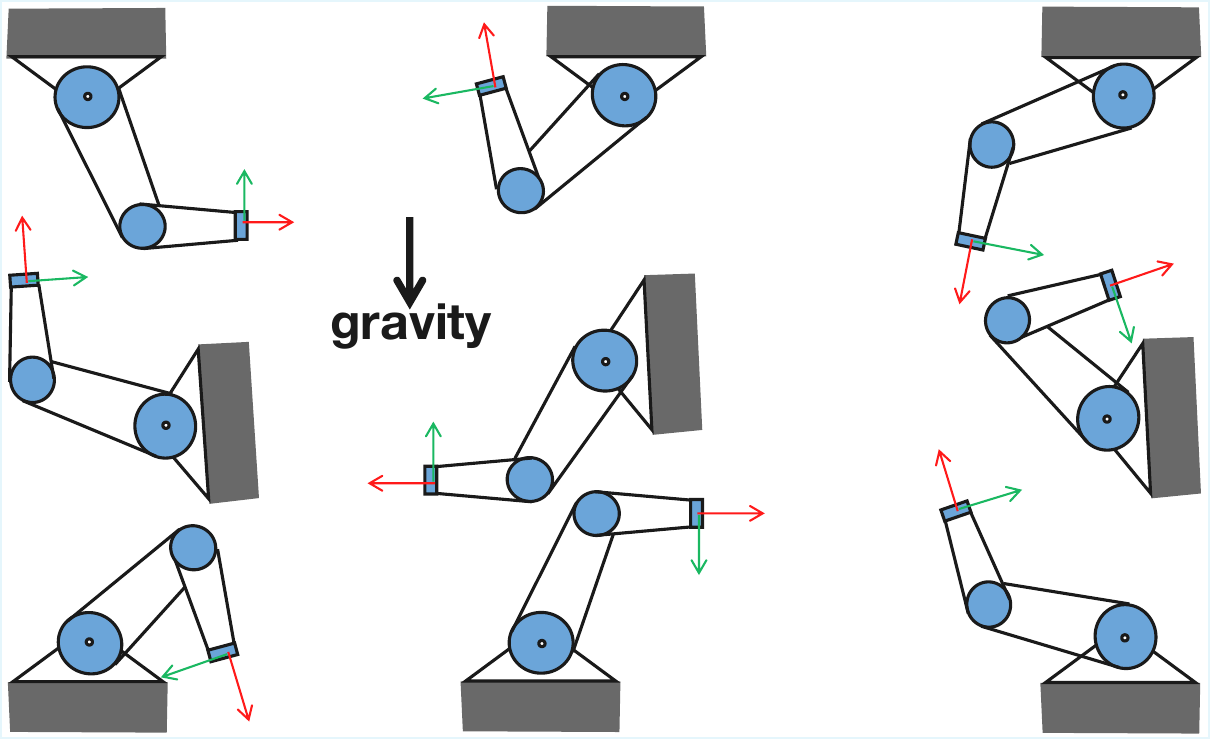}
    \end{minipage}
    \caption{
        Schematic of the attitude-aided accelerometer calibration.
        The method exploits external attitude information from a robotic arm, an IMU, a turntable, or other platforms.
        Regardless of sensor orientation or pose, the accelerometer always measures the inherent
        downward gravity vector, as illustrated for the IMU (left) and the robot (right).
    }
    \label{fig_wrist_mounted_sensor}
    \vspace{-0.2em}
\end{figure}

The main contributions of this work are summarized as follows:
\begin{itemize}
    \item A combined error matrix (CEM) is formulated to provide a unified representation
    of sensor errors, from which the scale factors, non-orthogonality,
    and alignment rotations matrix are analytically extracted.
    \item Based on the CEM, the proposed ALAC method for accelerometers provides a closed-form
    solution applicable to any attitude-available platform (e.g., turntables, robotic arms, or IMUs).
    \item ALAC achieves accurate and robust online calibration of IMU accelerometers using onboard attitude data.
\end{itemize}
The method is experimentally validated on two reference platforms: a robot arm and an IMU.
The paper is organized as follows:
Section~\ref{related_work} reviews related studies,
Section~\ref{section_model} establishes the calibration model,
Section~\ref{proposed_method} presents its closed-form solutions,
Section~\ref{section_exp} reports experimental validation and comparisons,
and Section~\ref{section_conclusion} concludes the paper.

\section{Related Work} \label{related_work}
\subsection{Reference-based Calibration}
Static multi-position methods are classical calibration approaches in which the sensor
is mounted on a high-precision turntable. By aligning sensor axes with the local vertical
direction, gravitational acceleration can be obtained~\cite{titterton2004_turntable, syed2007_turntable, fang2014_turntable, luczak2022_turntable, wang2023mems_turntable, grip2011_linear_bs}.
Although effective for estimating error parameters, their accuracy strongly depends on sensor alignment,
which is determined by the platform reference and orientation accuracy as well as the sensor installation accuracy.

To mitigate alignment limitations, \citet{Renk2005} employed a robot arm to acquire datasets,
where nonlinear alignment rotations were optimized using quasi-Newton minimization. Building on this
model, \citet{beravs2012} introduced an additional non-orthogonality error and applied the
unscented Kalman filter (UKF) to accelerate convergence by predicting next robot orientation.
Recently, \citet{khaula2025_robot} utilized a robot arm under small-angle alignment assumption
to estimate sensor bias by fixing it at six static orientations and then to formulate a linear
least-squares model.

For in-field calibration, \citet{dong2020_cuboid} designed a custom cuboid structure to decompose
parameters using twelve specially arranged attitudes, while \citet{xu2021novel} proposed a
six-position method based on arithmetic operations. \citet{duchi2023_tls_cuboid} introduced a linear
TLS approach that estimated all error coefficients except alignment rotation,
making its accuracy dependent on precise sensor-to-housing alignment.

Compared with these approaches, our ALAC method also leverages attitude information but
integrates all error factors within a unified linear formulation. This design avoids dependence on
specific sensor fixtures (e.g., strict alignment orientations or installation constraints) and
removes the need for iterative optimization.

Beyond the above methods, sensor redundancy has been explored.
\citet{belkhouche2019_2accel} used two oppositely mounted accelerometers for refinement,
and \citet{lu2022_fog_imu} employed a FOG-aided Kalman filter with near-parallel alignment.
Both approaches require accurate installation under small-error conditions.
Our CEM algorithm effectively alleviates these constraints when applied to redundant sensor configurations.

Another line of research leverages IMU-internal measurements for self-calibration, as discussed in
Section~\ref{section_imu_calibration}.

\subsection{IMU Accelerometer Calibration} \label{section_imu_calibration}
IMUs, integrating accelerometers, gyroscopes, and magnetometers, are widely used for attitude
estimation and motion tracking. Accelerometer calibration within IMU is often performed through
self-calibration methods, which eliminate the need for external equipment and enable convenient
in-field operation. The self-calibration methods mainly rely on iterative procedures such as
Newton's method~\cite{rohac2015_newton, lv2016_nl_tls_cuboid, qureshi2017_newton, li2019_newton},
the Levenberg-Marquardt (LM) algorithm~\cite{sipos2012, frosio2012_lm, tedaldi2014_lm},
gradient descent~\cite{lu2016_gradient_descent},
generalized nonlinear least squares~\cite{zou2024_gnls},
and meta-heuristic strategies including genetic algorithms (GA)~\cite{cui2017_ga}
and particle swarm optimization (PSO)~\cite{cai2013, poddar2019_pso, pesti2023_pso}.
Although iterative optimization can achieve accurate parameter estimates,
these methods often involve complex computations with Jacobians or Hessians,
and their convergence is not always guaranteed~\cite{ye2017_linearization}.
In~\cite{lv2016_nl_tls_cuboid}, intrinsic parameters were identified through iterative optimization
and then alignment rotations were estimated using 24 orientations generated by a cube-tray setup.
In the ellipsoid-fitting approach, Chao et al.~\cite{chao2022_ellipsoid}
utilized two fixed orientations to ensure a unique alignment rotation.

Several works~\cite{dong2020_cuboid, xu2021novel, duchi2023_tls_cuboid} developed reference-based
methods employing precision 3D-printed cuboids to simplify modeling with reference information.
\citet{troni2019_field} presented an online angular-rate-aided calibration method utilizing internal gyroscope
measurements to estimate sensor biases.
\citet{al2023_imu_hand} introduced a Kalman-filter-based hand-held calibration method
that jointly estimates accelerometer and gyroscope errors online by leveraging
velocity residuals and gravity constraints.
It assumes perfect alignment between the accelerometer triad and the IMU body frame,
and small-angle rotation between the navigation and world frames.
\citet{duchi2024_imu_tls} relaxed the small-angle assumption and extended the linear TLS framework
to an attitude-aided formulation, while retaining the perfect accelerometer-IMU alignment assumption.
In contrast, the proposed ALAC method removes both assumptions and, using homogeneous
linear equations in both offline or online modes,
estimates the transformations between the accelerometer, IMU body, navigation, and world frames.

\section{Prior Sensor and Calibration Model} \label{section_model}
\subsection{Accelerometer Sensor Error Model}
In MEMS-based accelerometers, raw sensor outputs (e.g., voltage or digital counts) are converted into physical
acceleration units. For triaxial accelerometers, ideally, the three sensing axes are mutually orthogonal and aligned with the
platform frame; however, assembly imperfections introduce small misalignments and scale factor
deviations. A unified triaxial model~\cite{David2007} is employed to represent these
errors, while higher-order effects such as temperature drift and nonlinearity are neglected.

\begin{figure}[h]
    \centering
    \begin{minipage}[t]{0.2\linewidth}
        \centering
        \includegraphics[width=0.7\linewidth]{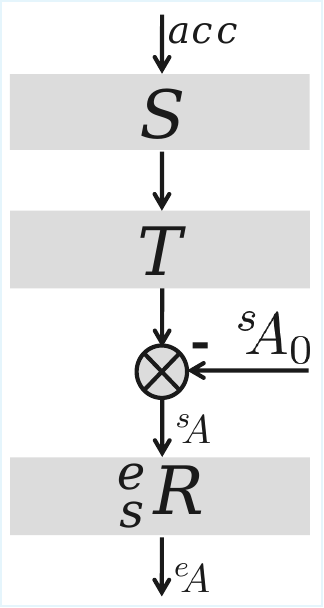}
        \caption{Accelerometer sensor model.}
        \label{fig_sensor_model}
    \end{minipage}
    \hfill
    \begin{minipage}[t]{0.3\linewidth}
        \centering
        \includegraphics[width=\linewidth]{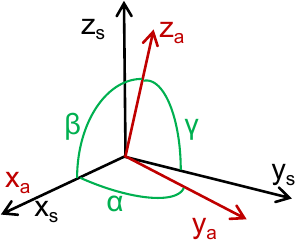}
        \caption{Gram--Schmidt Orthogonalization process.}
        \label{sensor_orthogonalization}
    \end{minipage}
    \hfill
    \begin{minipage}[t]{0.4\linewidth}
        \centering
        \includegraphics[width=\linewidth]{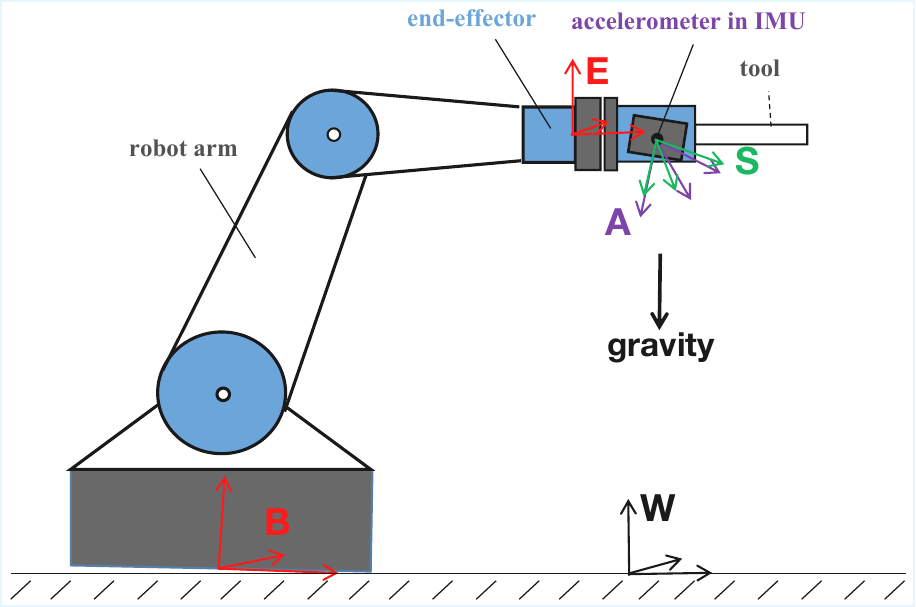}
        \caption{Robot-mounted accelerometer.}
        \label{fig_robot_accel}
    \end{minipage}
    \vspace{-0.8em}
\end{figure}

The physical accelerations $^e\!A$ in the external reference frame $e$ and raw accelerometer measurements $acc$ are
\begin{equation}
    {^e\!A} = [{^e\!A_x}; {^e\!A_y}; {^e\!A_z}], \quad
    acc = [acc_x; acc_y; acc_z] \in \mathbb{R}^{3 \times 1} .
\end{equation}
In order to obtain accurate acceleration quantities in the external reference frame $e$,
the raw sensor measurements $acc$ must be sequentially corrected.
This procedure, illustrated schematically in Fig.~\ref{fig_sensor_model}, involves four sequential steps:
\begin{enumerate}[noitemsep, topsep=0pt]
    \item \textbf{Scale factor correction}: apply the diagonal scale matrix $S \in \mathbb{R}^{3 \times 3}$;
    \item \textbf{Orthogonalization}: correct non-orthogonal axes to the sensor frame $s$ using $T \in \mathbb{R}^{3 \times 3}$;
    \item \textbf{Bias compensation}: subtract the bias ${^s\!A_0} \in \mathbb{R}^{3 \times 1}$ in the sensor frame $s$;
    \item \textbf{Alignment rotation}: rotate into the frame $e$ from $s$ with
    ${^e_sR} \in \mathbb{SO}(3)$ ($\{R \in \mathbb{R}^{3 \times 3} \mid
    R R^\top = R^\top R = I, \det R = 1\}$).
\end{enumerate}
Building on the above, the sensor model can be formulated as
\begin{equation}
    \label{equ_sensor_model}
    {^e\!A} = {^e_s\!R} \cdot ( T \cdot S \cdot acc - {^s\!A_0}) ,
\end{equation}
where
$
T = \begin{bmatrix}
 1   & 0   & 0  \\
 t_1 & 1   & 0  \\
 t_2 & t_3 & 1
\end{bmatrix}, \, t_1, t_2, t_3 \in [-1, 1];
\;
S = \begin{bmatrix}
 s_x & 0   & 0  \\
 0   & s_y & 0  \\
 0   & 0   & s_z
\end{bmatrix}, \, s_x, s_y, s_z > 0 .
$

As shown in \cite{David2007},
$T$ is first derived via the Gram-Schmidt algorithm (Fig.~\ref{sensor_orthogonalization}), and then approximated in a simplified form
since the inter-axis angles $\alpha$, $\beta$, $\gamma$ are typically close to $90^\circ$.

By incorporating the rotation ${^e_s\!R}$ into the transformation, we introduce a novel CEM to
reformulated the model as
\begin{equation}
    \label{equ_sensor_model_c}
    {^e\!A} = C \cdot acc - {^e\!A_0} ,
\end{equation}
where $C={^e_sR} \cdot T \cdot S$ is the CEM and
${^e\!A_0}={^e_sR} \cdot {^s\!A_0}$ is the bias expressed in the external reference frame.
The essential task of this work is to extract ${^e_sR}$, $T$, and $S$ from the
CEM $C$, as detailed in Section~\ref{sec_extraction}.
Since $C$ is identified as a general $3 \times 3$ matrix from data,
the parameterizations of $T$ and $S$ can be cast into a unified model, within which the prior
approximations remain valid without compromising accuracy.

\subsection{Attitude-Aided Calibration Model}
Following the accelerometer model, the external reference frame $e$ is defined as the frame for which
attitude information is available. Depending on the experimental setup, $e$ may represent
the robot end-effector frame, the IMU body frame, or another external platform providing
orientation (e.g., a turntable system). The rotation matrix ${^e_sR}$ denotes the alignment
from the sensor frame $s$ to this chosen external frame $e$.

Accelerometer measurements inherently include the projection of gravity onto the sensor axes.
In the case of a robot manipulator with a wrist-mounted accelerometer (Fig.~\ref{fig_robot_accel}),
this gravitational component can be compensated using the robot orientation. The gravitational accelerations in the end-effector
frame $e$ are formulated as
\begin{equation}
    \label{equ_robot_end_acceleration}
    ^e\!A_\text{true} = {^e\!A} - {_{b}^{e}R} \cdot {^{b}_w\!R} \cdot {^w\!A_{g}},
\end{equation}
where $^e\!A_\text{true}$ denotes the true accelerations of the end-effector expressed in $e$,
$^e\!A$ are the corrected accelerometer measurements, ${_{b}^{e}R} \in \mathbb{SO}(3)$ is the
forward-kinematics orientation from the robot base $b$ to the end-effector $e$,
${_{w}^{b}\!R} \in \mathbb{SO}(3)$ is the rotation of the world frame $w$ with respect to $b$, and
$^{w}\!A_{g} = [0; 0; g]_{3\times 1}$ is the local gravity vector in $w$ ($g = 9.80665\:\mathrm{m/s^2}$).
For IMU, $e$ represents the IMU body frame, and $b$ represents the navigation frame.

To simplify \eqref{equ_robot_end_acceleration} and avoid explicit dependence on both ${_{w}^{b}R}$
and ${^w\!A_g}$, an auxiliary variable is introduced as
\begin{equation}
    \label{equ_A_b}
    ^b\!A_g = {_{w}^{b}R} \cdot {^w\!A_g}, \quad \text{s.t. } \|{^b\!A_g}\|_2 = g .
\end{equation}

Substituting \eqref{equ_sensor_model_c} and \eqref{equ_A_b} into \eqref{equ_robot_end_acceleration},
the acceleration of robot end can be rewritten as
\begin{equation}
    \label{equ_robot_end_acceleration_v1}
    ^e\!A_\text{true} = C \cdot acc - {^e\!A_0} - {_{b}^{e}R} \cdot {^b\!A_g}
\end{equation}
When the robot is stationary, the accelerometer is static with respect to the environment,
and the true end-effector acceleration satisfies $^e\!A_\text{true}=0$. Under this condition, the sensor
output reflects only the gravitational acceleration, bounded by $\pm g$. Exploiting this property,
the calibration model is expressed as
\begin{equation}
    \label{equ_calibration_model_v0}
    C \cdot acc - {^e\!A_0} - {_{b}^{e}R} \cdot {^b\!A_g} = 0,
    \quad \text{s.t. } \|{^b\!A_g}\|_2 = g ,
\end{equation}
where the available measurements are the raw accelerometer output $acc$ and the orientation
${_{b}^{e}R}$.

In this compact representation, the entries of $C$, together with ${^e\!A_0}$ and ${^b\!A_g}$,
are grouped into the parameter vector $x \in \mathbb{R}^{15 \times 1}$ (see \eqref{equ_x}).
The parameters are identified by solving a constrained homogeneous least-squares (CHLS) problem (Section~\ref{sec_cls_identification}),
from which ${^e_sR}$, $T$, $S$, ${^s\!A_0}$, and ${^b\!A_g}$ are analytically recovered (Section~\ref{sec_extraction}).
This unified formulation, which forms the theoretical basis of the proposed ALAC method,
is applicable across different platforms.
For instance, when $e$ denotes the IMU body frame and $b$ the navigation frame,
${^e_bR}$ represents the orientation of the navigation frame relative to the IMU body frame,
and ${^b\!A_g}$ is the gravity vector expressed in the navigation frame.

\section{Proposed Method} \label{proposed_method}
\subsection{Constrained Least-Squares Calibration} \label{sec_cls_identification}
As established in \eqref{equ_calibration_model_v0}, the proposed ALAC formulation involves estimating
the CEM $C$, the transformed bias ${^e\!A_0}$, and the gravity vector ${^b\!A_g}$.
For a compact formulation, these unknowns are collected into a single parameter vector
$x \in \mathbb{R}^{15 \times 1}$ defined in \eqref{equ_x}. This representation allows the calibration
model to be cast as a CHLS problem, which is solved efficiently using standard algebra methods.

The unknown parameters are collected into the vector $x \in \mathbb{R}^{15 \times 1}$, defined as
\begin{equation}
    \label{equ_x}
    x = [c_{11}; c_{12}; \ldots; c_{33}; {^e\!A_0}; {^b\!A_g}] ,
\end{equation}
where $c_{ij}$ are the entries of the CEM $C$. Using this representation, the
calibration model in \eqref{equ_calibration_model_v0} can be expressed as the homogeneous system
\begin{equation}
    \label{equ_AX_definision}
    Ax = 0 \quad \text{s.t.} \: \|Dx\|_2 = 1 ,
\end{equation}
with
$
\begin{aligned}
    A &= \left[\left[\begin{array}{ccc}
    acc^\top & 0_{1\times 3} & 0_{1\times 3} \\
    0_{1\times 3} & acc^\top & 0_{1\times 3} \\
    0_{1\times 3} & 0_{1\times 3} & acc^\top
    \end{array}\right],-I_{3\times 3},-{_{b}^{e}R}_{3\times 3}\right] \in \mathbb{R}^{3n \times 15},
    D &=\begin{bmatrix}
        0_{12\times 12} & 0_{12\times 3}\\
        0_{3\times 12} & \frac{I_{3\times 3}}{g}
    \end{bmatrix} .
\end{aligned}
$

Given $n$ reference orientations ${_{b}^{e}R}_{i}$ ($i=1,2,\ldots,n$, $n>4$)
and their corresponding raw accelerometer measurements $acc_i$, define
the per-pose matrix $A_i$ and the stacked block matrix $\bar A$ as
\begin{equation}
    \label{eq:A_and_Abar}
    \begin{aligned}
        A_i &=
        \left[
        \left[
        \begin{array}{ccc}
            acc_i^\top & 0_{1\times 3} & 0_{1\times 3} \\
            0_{1\times 3} & acc_i^\top & 0_{1\times 3} \\
            0_{1\times 3} & 0_{1\times 3} & acc_i^\top
        \end{array}
        \right] ,
        -I_{3\times 3},\ -{_{b}^{e}R}_{i}
        \right] \in \mathbb{R}^{3\times 15},
        \qquad
        \bar{A} \triangleq
        \begin{bmatrix}
            A_1 \\ \vdots \\ A_n
        \end{bmatrix} \in \mathbb{R}^{3n \times 15}.
    \end{aligned}
\end{equation}
The stacked formulation leads to the compact CHLS problem
\begin{equation}
    \label{equ_J}
    \min_{x} \|\bar{A}x\|_2^2 \quad \text{s.t. } \|Dx\|_2 = 1 .
\end{equation}
In ALAC, this CHLS formulation is efficiently solved using established algebraic methods such as
singular value decomposition (SVD) or generalized eigenvalue problem
(GEVP)~\cite{GolubVanLoan2013,BoydVandenberghe2004}. Unlike iterative optimization
approaches such as Newton or Levenberg-Marquardt, these algebraic methods provide a direct
computation of the solution.
The detailed GEVP formulation is provided in Appendix~\ref{app_gevp}, and its reduction to a
standard eigenvalue problem (SEVP) exploiting the block-diagonal structure of $D$ is presented in
Appendix~\ref{app_sevp}. A recursive GEVP-based extension for online calibration is described in
Appendix~\ref{app_recursive}.

Let $x^*$ denote the optimal solution of \eqref{equ_J}. The estimate is then
\begin{equation}
    \label{equ_x_estimation}
    \hat{x} = \pm x^* ,
\end{equation}
where the symbol $\hat{}$ denotes the estimate of the corresponding parameter.
The sign ambiguity is inherent to homogeneous formulations. This ambiguity is resolved in the
parameter extraction stage (see Section~\ref{sec_extraction}), since only one choice yields a valid
rotation matrix ${^e_sR}$, thereby uniquely determining the final estimate.

\subsection{Parameters Extraction} \label{sec_extraction}
Having identified the parameter vector $x$ in Section~\ref{sec_cls_identification},
the next step in ALAC is to extract the calibration parameters in \eqref{equ_calibration_model_v0}
from its estimate $\hat{x}$ given in \eqref{equ_x_estimation}.
We assume $\hat{x}=x^*$ first, then decide its sign by valid rotation matrix.

Since the first nine entries of $x$ correspond to the CEM $C$ (see \eqref{equ_x}), the estimate
$\hat{C}$ is obtained as
\begin{equation}
    \label{equ_C_estimation}
    \hat{C} =
    \begin{bmatrix}
        x^*_{1} & x^*_{2} & x^*_{3} \\
        x^*_{4} & x^*_{5} & x^*_{6} \\
        x^*_{7} & x^*_{8} & x^*_{9}
    \end{bmatrix}.
\end{equation}
As defined in \eqref{equ_sensor_model_c}, the CEM is formulated as
$C = {^e_sR} \cdot T \cdot S$, which inherently yields nine scalar equations.
Here, $S$ is a diagonal sensitivity matrix, $T$ is a lower-triangular nonorthogonality matrix,
and ${^e_sR} \in \mathbb{SO}(3)$ is a rotation matrix.
These structural constraints ensure that ${^e_sR}$, $T$, and $S$ are uniquely
determined from the estimated $\hat{C}$.

From \eqref{equ_sensor_model}, the $\hat{C} = {^e_sR}\cdot T\cdot S$ can be decomposed and
explicitly written as
\begin{equation}
    \label{equ_RST_estimation}
    \begin{aligned}
    \left[ \hat{c}_{1}, \hat{c}_{2}, \hat{c}_{3} \right]
    &=\left[r_{1}, r_{2}, r_{3}\right]\left[
        \begin{array}{ccc}
        1 & 0 & 0 \\
        t_{1} & 1 & 0 \\
        t_{2} & t_{3} & 1
        \end{array}
    \right]\left[\begin{array}{ccc}
        s_{x} & 0 & 0 \\
        0 & s_{y} & 0 \\
        0 & 0 & s_{z}
    \end{array}\right] \\
    &=\begin{bmatrix}
        s_{x} r_{1}+s_{x} t_{1} r_{2}+s_{x} t_{2} r_{3}, & s_{y} r_{2}+s_{y} t_{3} r_{3}, & s_{z} r_{3}
    \end{bmatrix} ,
    \end{aligned}
\end{equation}
where $r_i$ and $\hat{c}_i$ ($i=1, 2, 3$) denoting the column vectors of ${^e_sR}$ and $\hat{C}$, respectively.

Firstly, from \eqref{equ_RST_estimation}, we obtain
\begin{equation}
    \hat{c}_{3} = s_{z} r_{3}.
\end{equation}
Accordingly, the following estimations hold:
\begin{equation}
    \begin{aligned}
        \label{equ_sz_equ_r3}
        \hat{s}_{z} &= \left|\hat{c}_{3}\right| \; (>0), \quad
        \hat{r}_{3} = \hat{c}_{3}/\left|\hat{c}_{3}\right|.
    \end{aligned}
\end{equation}

Next, taking the cross product of $\hat r_{3}$ with second column
$\hat c_{2}=s_{y}r_{2}+s_{y}t_{3}\hat r_{3}$, the second scale of $S$ is determined as
\begin{equation}
    \hat s_{y}=|\hat c_{2}\times\hat r_{3}|/|r_{2}\times\hat r_{3}|=|\hat c_{2}\times\hat r_{3}|\;\;(>0).
\end{equation}
Similarly, $\hat{r}^\top_{3}$ is post-multiplied by $\hat{c}_2 = \hat{s}_{y} r_{2}+\hat{s}_{y} t_{3} \hat{r}_{3}$,
and the third scale of the $T$ is given by
\begin{equation}
    \label{equ_t3}
    \hat{t}_{3} = \hat{r}^\top_{3}\cdot \hat{c}_{2}/\hat{s}_{y}.
\end{equation}
Since $\hat{s}_{y}$ is nonzero, the second column of the
rotation matrix $^e_sR$ can be estimated by
\begin{equation}
    \label{equ_r2}
    \hat{r}_{2} = \hat{c}_{2}/\hat{s}_{y} - \hat{t}_{3}\cdot \hat{r}_3.
\end{equation}

In addition, by summing the cross products between the first column $\hat{c}_1=s_{x} r_{1}+s_{x} t_{1} r_{2}+s_{x} t_{2} r_{3}$ of
\eqref{equ_RST_estimation} and the estimated vectors $\hat{r}_{2}$ and $\hat{r}_{3}$,
the first scale of $S$ is
\begin{equation}
    \label{equ_sx}
    \hat{s}_{x} = \frac{\left|\hat{c}_{1} \times \hat{r}_{2} + \hat{c}_{1} \times \hat{r}_{3}\right|}{\left|\hat{r}_{3} - \hat{r}_{2}\right|} \;\; (>0).
\end{equation}
Left-multiplying the first column $\hat{c}_1=\hat{s}_{x}(r_{1}+t_{1}r_{2}+t_{2}r_{3})$  by $\hat{r}^\top_{2}$ leads to the first scale of $T$:
\begin{equation}
    \label{equ_t1}
    \hat{t}_{1} = \hat{r}^\top_{2}\cdot \hat{c}_{1}/\hat{s}_{x}.
\end{equation}
Similarly, projecting onto $\hat{r}^\top_{3}$ yields
\begin{equation}
    \label{equ_t2}
    \hat{t}_{2} = \hat{r}^\top_{3}\cdot \hat{c}_{1}/\hat{s}_{x}.
\end{equation}
Hence, under the assumption $\hat{s}_{x} > 0$, the first column of the rotation matrix $^e_sR$
can be expressed as
\begin{equation}
    \label{equ_r1}
    \hat{r}_{1} = \hat{c}_{1}/\hat{s}_{x} - \hat{t}_{1}\cdot \hat{r}_2 - \hat{t}_{2}\cdot \hat{r}_3.
\end{equation}

Bringing these results together, the estimates of $\hat{S}$, $\hat{T}$ and a matrix
$R_1 = [\hat{r}_1, \hat{r}_2, \hat{r}_3]$ are obtained.
It get another matrix $R_2 = -R_1$, same $\hat{S}$ and $\hat{T}$
that we substitute $\hat{x}=x^*$ into \eqref{equ_C_estimation}-\eqref{equ_r1}.
Because of $det\; R_1 = -det\; R_2$, only one of them is equivalent to $+1$, so that
the sign of $\hat{x}$ in \eqref{equ_x_estimation} is fixed and $^e_sR \in \mathbb{SO}(3)$ is identified.

Now, the gravity and bias vector can be estimated from the determined $\hat{x}$.
By \eqref{equ_x}, the gravitational vector $^b\!A_g$ is
\begin{equation}
    \label{equ_bAg}
    \hat{^b\!A_g} = \hat{x}_{13-15}.
\end{equation}

Finally, combining \eqref{equ_x} and \eqref{equ_sensor_model_c}, the bias in the sensor
frame is obtained in two steps: first estimating $\hat{^e\!A}_{0}$ from $\hat{x}_{10-12}$, and then
transforming it into the sensor frame as
\begin{equation}
    \begin{aligned}
        \label{equ_bias}
        \hat{^e\!A_0} &= \hat{x}_{10-12}, \quad
        \hat{^s\!A_0} = \hat{^e_s\!R^\top} \cdot \hat{^e\!A_0}.
    \end{aligned}
\end{equation}

Thus, the proposed ALAC procedure consistently identifies all unknown variables, namely
$^e_sR$, $T$, $S$, $^s\!A_0$, and $^b\!A_g$,
from accelerometer measurements and external orientation information.

\section{Experimental Validation} \label{section_exp}
This section evaluates the proposed ALAC method using synthetic data, a real robot-mounted IMU
accelerometer, and an open-source IMU dataset.
Unless specified otherwise, the notation follows Section~\ref{proposed_method}.

\subsection{Evaluation Protocol and Metrics}
To evaluate calibration quality, we report three complementary metrics over
the test set $\mathcal{T}$ of size $N$:
\begin{itemize}
    \item \textbf{Estimation error} $\epsilon_\text{est}$:
    Quantifies the overall discrepancy between estimated and referenced parameters:
    \[
    \epsilon_\text{est} = \sum
    \Big( \|p_{j,\text{est}} - p_{j,\text{ref}}\|_2 \;\; (p_{j,\text{est}} \in \mathbb{R}^3), \quad
        \|p_{j,\text{est}} - p_{j,\text{ref}}\|_F \;\; (p_{j,\text{est}} \in \mathbb{R}^{3\times 3}) \Big).
    \]

    \item \textbf{Root Mean Square Error (RMSE)} $\epsilon_\text{rmse}$:
    In static or quasi-static conditions, evaluates whether the calibrated accelerometer output has the correct norm with respect to
    the local gravity magnitude $g$:
    \[
    \epsilon_{\text{rmse}} =
    \sqrt{\frac{1}{N} \sum\nolimits_{i=1}^{N}
    \Big(\,\|\hat{T}\cdot \hat{S} \cdot acc_i - \hat{^{s}\!A}_0\|_2 - g \Big)^2} \quad (\mathrm{m/s^2}).
    \]

    \item \textbf{Compensation error} $\epsilon_\text{comp}$:
    Evaluates the residual acceleration after the correction and gravity compensation
    ($^s\!A_\text{true}=0$ in static or quasi-static conditions), affected by the
    precision of $acc_i$ and $_b^e\!R_i$:
    \[
        \epsilon_\text{comp} = \frac{1}{N} \sum\nolimits_{i=1}^{N}
        \Big\|
        {^s\!A_\text{true}} -
        (\hat{T} \cdot \hat{S} \cdot acc_i - \hat{^{s}\!A}_0
        - {^s_e\hat{R}} \cdot {_{b}^{e}R}_i \cdot \hat{^{b}\!A_g})
        \Big\|_2 \quad (\mathrm{m/s^2}).
    \]
\end{itemize}

Note that $\epsilon_\text{rmse}$ measures only the consistency
between the accelerometer norm and gravity magnitude, and thus is often smaller.
By contrast, $\epsilon_\text{comp}$ additionally depends on external orientation references,
making it more sensitive to modeling errors and noise, and therefore potentially larger
unless the external reference is highly accurate.

\subsection{Numerical Experiments} \label{sec_simulation}
\paragraph{Data generation.}
Synthetic datasets were generated by sampling $n$ random orientations ${_{b}^{e}R}_i$ and
random accelerations $^s\!A_\text{true}$ (zero for calibration).
Accelerometer measurements were then synthesized as:
\begin{equation}
    acc_i = S^{-1} T^{-1}({^s\!A_\text{true}} + {_e^sR}\cdot {_b^eR}_i\cdot {^b\!A_g} + {^s\!A_0}),
    \quad i=1,\ldots,n \; (n>4),
\end{equation}
optionally perturbed by i.i.d. Gaussian noise
$\eta_i\sim\mathcal{N}(0,\sigma^2 I_3)$ with $\sigma\in[0,\,0.01]$ m/s$^2$.
The sampled orientations ${_{b}^{e}R}_i$ were assumed to be noise-free.

\paragraph{Parameter setup.}
Six representative configurations were evaluated. Each setup specifies
$S=\mathrm{diag}(s_x,s_y,s_z)$, $T=I+\mathrm{tril}(t_1,t_2,t_3)$,
${_{s}^{e}\!R}=R_Z(\alpha)R_Y(\beta)R_X(\gamma)$ ($\alpha$, $\beta$, $\gamma$ in degrees),
$^{b}\!A_g$, and $^{s}\!A_0$.

\paragraph{Procedures and results.}
For each setup we: (i) set ${^s\!A_\text{true}}=0$ and $n=24$ to generate a dataset for calibration;
(ii) estimated the unknown parameters using ALAC and compute $\epsilon_\text{est}$;
(iii) evaluated on an independent test set ($n=150$ and random ${^s\!A_\text{true}}$) to obtain
$\epsilon_\text{comp}$ and $\epsilon_\text{rmse}$.
As summarized in Table~\ref{tab:num_setups}, the algorithm works well when all three errors remain close to the injected
Gaussian noise level ($\sigma=0.01$ m/s$^2$) across the six setups.
\begin{table}[h]
\vspace{-0.5em}
\centering
\caption{Parameter setups.}
    \label{tab:num_setups}
    \small
    \begin{tabular}{cccccc}
    \toprule
    Setup & $[s_x, s_y, s_z]$ & $[t_1, t_2, t_3]$ & $[\alpha,\beta,\gamma]$ & $^{b}\!A_g$ (m/s$^2$) & $^{s}\!A_0$ (m/s$^2$)\\
    \midrule
    1 & $[1,1,1]$ & $[0,0,0]$ & $[0,7,-3]$ & $[0;0;-9.81]$ & $[0;0;0]$ \\
    2 & $[1,1,1]$ & $[0,0,0]$ & $[0,-7,3]$ & $[0;0;9.81]$ & $[3;1;-4]$ \\
    3 & $[1,1,1]$ & $[0,0,0]$ & $[0,7,3]$ & $[1;-2;9.55]$ & $[3;-1;4]$ \\
    4 & $[1,1,1]$ & $[0,0,0]$ & $[45,15,-13]$ & $[-1;2;9.55]$ & $[-3;1;4]$ \\
    5 & $[1,1,1]$ & $[0.05,0.08,0.03]$ & $[45,-15,13]$ & $[1;2;9.55]$ & $[3;1;4]$ \\
    6 & $[0.9,1.3,0.8]$ & $[0.05,0.08,0.03]$ & $[-45,15,13]$ & $[1;2;-9.55]$ & $[-3;1;-4]$ \\
    \bottomrule
    \end{tabular}
    \vspace{-0.8em}
\end{table}

\subsection{Robot-mounted Accelerometer Experiments}
\begin{figure}[htbp]
    \vspace{-0.55em}
    \centering
    \begin{minipage}[b]{0.41\textwidth}
        \vspace{0pt}
        \centering
        \includegraphics[width=\linewidth]{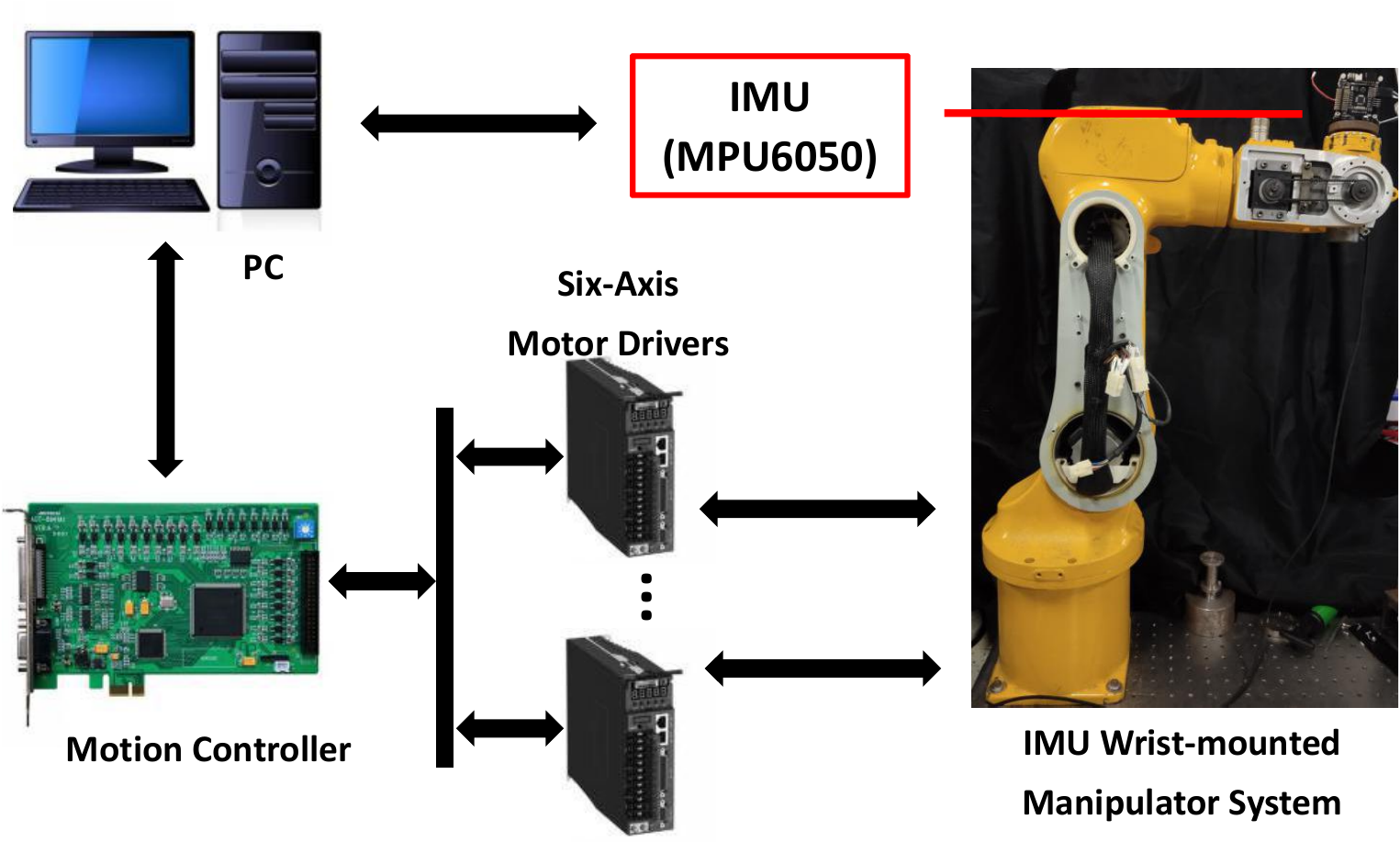}
        \caption{Robot-mounted IMU.}
        \label{fig_acce_expSetup}
    \end{minipage}
    \hfill
    \begin{minipage}[b]{0.57\textwidth}
        \vspace{0pt}
        \begin{minipage}[t]{\linewidth}
            \vspace{0pt}
            \centering
            \captionsetup{type=table}
            \caption{Calibration errors of different algebraic methods.}
            \label{tab_diff_error}
            \vspace{-1.0ex}
            \begin{tabular}{ccccc}
                \toprule
                method & SVD & GEVP & SEVP & Recursive GEVP \\
                \midrule
                $\epsilon_\text{comp}$ & 0.33808 & 0.33808 & 0.33808 & 0.33808 \\
                \midrule
                $\epsilon_\text{rmse}$ & 0.00913 & 0.00913 & 0.00913 & 0.00913 \\
                \bottomrule
            \end{tabular}
        \end{minipage}
        \begin{minipage}[b]{\linewidth}
            \centering
            \includegraphics[width=\linewidth]{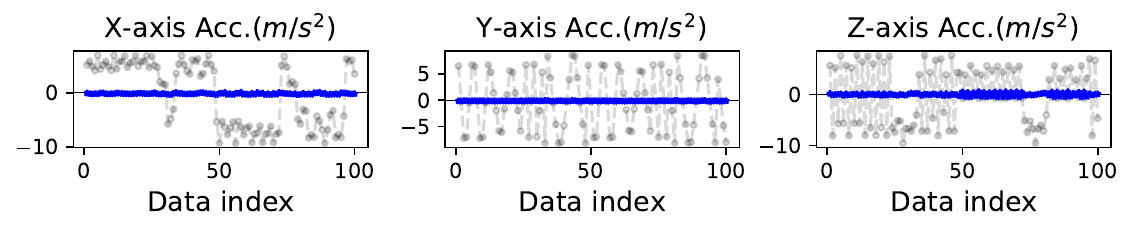}
            \captionof{figure}{Validation results (blue solid) and raw data (gray dashed).}
            \label{fig_acce_exp_val}
        \end{minipage}
    \end{minipage}
\end{figure}

\paragraph{Platform and data.}
A commercial IMU (MPU6050) was securely mounted on the wrist of an EFFORT ER3B-C20 manipulator,
as shown in Fig.~\ref{fig_acce_expSetup}, and all signals were continuously sampled at a frequency
of 400~Hz. Under static conditions, measurements were collected over fixed time interval and
subsequently averaged to obtain a representative value. A total of 216 static points were acquired
throughout the full workspace.
\begin{table}[h]
    \vspace{-1.2em}
	\caption{\label{tab_acce_idenPara}Estimated parameters.}
    \centering
	\resizebox{0.9\textwidth}{10mm}{
	\begin{tabular}{ccccc}
	\toprule
		{$S$} & {$T$} & {$^s_eR$} & {$^b\!A_g$ {($\mathrm{m/s^2}$)}} & {$^s\!A_0$ {($\mathrm{m/s^2}$)}} \\

	\midrule
		$\begin{bmatrix}
			0.9936 & 0 & 0 \\ 0 & 1.0051 & 0 \\ 0 & 0 & 0.9964
		\end{bmatrix}$
		&$\begin{bmatrix}
			1 & 0 & 0 \\ 0.0004 & 1 & 0 \\ 0.0117 & -0.0022 & 1
		\end{bmatrix}$
		& $\begin{bmatrix}
			-0.8613 & -0.5081 & -0.0077 \\
			-0.0037 & 0.0213 & -0.9998 \\
			0.5081 & -0.8610 & -0.0203
		\end{bmatrix}$
		& $\begin{bmatrix}
			-0.0230 \\ -0.0062 \\ 9.7888
		\end{bmatrix}$
		& $\begin{bmatrix}
			0.3579 \\ -0.2209 \\ 0.2385
		\end{bmatrix}$ \\
	\bottomrule
	\end{tabular}}
    \vspace{-1.0em}
\end{table}
\paragraph{Comparison of algebraic methods.}
With the 24 static orientations for estimation and 100 for evaluation, we used SVD, GEVP,
SEVP (in Appendix~\ref{app_sevp}), Recursive GEVP (in Appendix~\ref{app_recursive}) to solve
the CHLS problem in \eqref{equ_J}.
As illustrated in Table~\ref{tab_diff_error}, all methods
yield identical results up to the 5 decimal places reported; in our experiments, discrepancies
only appear beyond the 15th decimal place, i.e., at the level of machine precision.
We recommend GEVP for its simplicity and established reliability.

\paragraph{Calibration performance.}
Using the 24 static orientations for estimating and 100 for evaluating,
the Table~\ref{tab_acce_idenPara} shows the estimated parameters, and Fig.~\ref{fig_acce_exp_val}
indicates that $^s\!A_x$, $^s\!A_y$, $^s\!A_z$ (bounded by $\pm g$) are compensated to $\pm 0.277$, $\pm 0.146$,
$\pm 0.509 \mathrm{m/s^2}$, respectively.
Table~\ref{tab_diff_error} reports $\epsilon_\text{comp}=0.33808$ and $\epsilon_\text{rmse}=0.00913$,
indicating small accelerometer uncertainty and large robotic uncertainty, and the latter of which is
effectively mitigated by our method.

\begin{figure}[htbp]
    \vspace{-0.95em}
    \centering
    \begin{minipage}[b]{0.63\textwidth}
        \vspace{0pt}
        \centering
        \includegraphics[width=\linewidth]{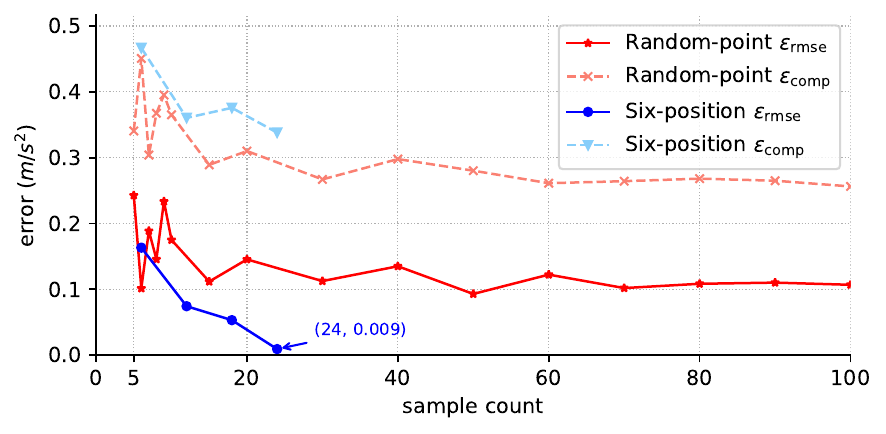}
        \caption{Effect of sampling strategy and sample count.}
        \label{fig_acce_pose_count_effect}
    \end{minipage}
    \hfil
    \begin{minipage}[b]{0.35\textwidth}
        \vspace{0pt}
        \centering
        \includegraphics[width=\linewidth]{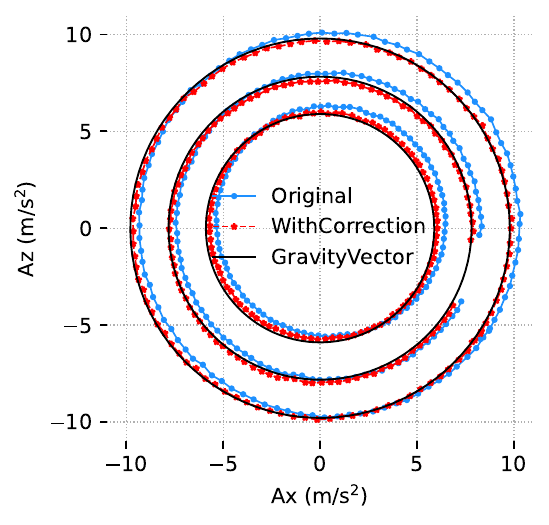}
        \caption{Accelerometer Correction.}
        \label{fig_acce_whyCorr}
    \end{minipage}
    \vspace{-1.5em}
\end{figure}

\paragraph{Sampling Strategy Comparison.}
Fig.~\ref{fig_acce_pose_count_effect} illustrates the effect of static-sample count $n$.
For random sampling, both $\epsilon_\text{rmse}$ and $\epsilon_\text{comp}$
drop initially but fluctuate for small $n$, and then level off around $n \approx 30$.
The six-position strategy is analogous to that of \cite{xu2021novel} and employs 24 orientations
(4 groups of 6 positions) as detailed in Appendix~\ref{app_24_orientations}. It converges faster:
errors decrease smoothly and $\epsilon_\text{rmse}$ approaches a stable lower bound around $n=24$.
Overall, six-position orientations coverage is more effective than adding random samples,
since the chosen orientations strongly excite the sensor axes and improve numerical
conditioning~\cite{xu2021novel}, and it can significantly overcome the uncertainty of orientations.
It is also observed that adding random samples to the six-position strategy does not further improve accuracy.

\paragraph{Accelerometer correction.}
To evaluate the correction effect, we designed a dynamic trajectory in which only the sixth joint of
the manipulator rotated at very low speed. This setup minimized acceleration along one axis,
thereby maintaining plane visibility.
As shown in Fig.~\ref{fig_acce_whyCorr}, the uncorrected accelerometer readings
(blue line with circle markers) deviate from the gravity sphere (black solid line), whereas the
corrected data (red line with star markers) match the sphere closely. It confirms that
the proposed method improves the accuracy and stability of the accelerometer output.

\paragraph{Baseline comparison on static data.}
Most reference-based methods explicitly assume ideal sensor alignment,
which makes them unsuitable for direct comparison.
To achieve a more comprehensive evaluation, a TLS method~\cite{duchi2023_tls_cuboid,duchi2024_imu_tls}
was included, as it disregards sensor installation errors.
In addition, the quasi-Newton method~\cite{Renk2005} and the UKF method~\cite{beravs2012} were
considered, both accounting for all relevant factors through nonlinear modeling.
Self-calibration methods are typically offline and formulated as iterative minimization of
$T\cdot S\cdot acc-{^s\!A_0}$.
For reference, we implemented bounded optimizers including (i) quasi-Newton~\cite{qureshi2017_newton},
(ii) LM~\cite{tedaldi2014_lm}, (iii) PSO,
and (iv) exact trust-region (TR-Exact) method. All were evaluated using the cost function
$\sum\nolimits_{i=1}^{N} \left( \| a_{\mathrm{cal}} \|^2_2 - \| g \|^2 \right)^2$,
with explicit Jacobians (and Hessian for TR-Exact).
For completeness of comparison, we further included the classical ellipsoid fitting approach,
which also follows the same formulation $T\cdot S\cdot acc-{^s\!A_0}$.
\begin{table}[t]
    \centering
    \small 
    \caption{
        Static-condition calibration errors for ALAC and baselines.
        Newton's/LM initialized by our estimate;
        Raw-data RMSE before calibration: 0.3118; $\epsilon_\text{est}$ relative to Ours-Six-position (reference).
        $^{\dagger}$TLS excludes ${^{e}_{s}\!R}$.
    }
    \label{tab_static_comparison}
    \setlength{\tabcolsep}{3.4pt}
    \renewcommand{\arraystretch}{1.1}
    \begin{tabular}{lccccccccc}
        \toprule
        & \multicolumn{5}{c}{Reference-based ($T$, $S$, ${^s\!A_0}$, ${^b\!A_g}$, ${^e_sR}$)$^{\dagger}$} & \multicolumn{4}{c}{Self-calibration ($T$, $S$, ${^s\!A_0}$)} \\
        \cmidrule(lr){2-6} \cmidrule(lr){7-10}
        & \textbf{Ours}-Six-position & \textbf{Ours}-Random & Newton's~\cite{Renk2005} & UKF~\cite{beravs2012} & TLS~\cite{duchi2023_tls_cuboid,duchi2024_imu_tls} & Newton's & LM & PSO & \textbf{Ellipsoid} \\
        \midrule
        $\epsilon_\text{rmse}$ & 0.0091 & 0.1097 & 0.1085 & 0.1219 & 4.022 & 0.0034 & 0.0034 & 0.0070 & 0.0035 \\
        $\epsilon_\text{comp}$ & 0.338 & 0.255 & 0.255 & 0.347 & 7.778& - & - & - & - \\
        $\epsilon_\text{est}$ & ref. & 0.238 & 25.48 & 0.279 & 22.03 & 0.013 & 0.013 & 14.02 & 0.014 \\
        time (s) & 8e-4 & 7e-4 & 3e-1 & 2e-1 & 2e-4 & 4e-2 & 2e-3 & 2.5 & 1e-4 \\
        data n & 24 & 24 & 24 & 132 & 24 & 24 & 24 & 24 & 24 \\
        sensitive & - & - & yes & unstable & - & yes & yes & yes & - \\
        \bottomrule
    \end{tabular}
\end{table}

Table~\ref{tab_static_comparison} compares the proposed and baseline methods,
indicating that our ALAC method significantly outperforms all reference-based baselines.
Neglecting installation errors makes the TLS method prone to large errors, rendering it unsuitable
for practical mechanism. Among the other two nonlinear estimators,
the quasi-Newton method~\cite{Renk2005} suffers from convergence instability, whereas the
UKF method~\cite{beravs2012} is overly sensitive to initialization and noise, both resulting in
larger errors than ours. For self-calibration methods, the PSO optimizer fails to converge
reliably due to the lack of an appropriate initialization mechanism. The remaining optimizers
(Newton's, LM, and Ellipsoid) achieve considerably lower RMSE, as self-calibration is only
affected by accelerometer uncertainty. Importantly, the RMSE of our method using 24 six-position orientations
remains on the same order as that of the self-calibration methods, highlighting its robustness
under attitude uncertainties (large $\epsilon_\text{comp}$).

\subsection{Quasi-static IMU Calibration}
\paragraph{Data.}
The experiments used the MPU6050RM3100 IMU dataset from the open-source repository~\cite{imu_dataset},
recorded during the eV50p (150 mm/s) motion. The dataset provides synchronized
accelerometer and gyroscope measurements along controlled robot trajectories.
The attitude was independently estimated using separate sensor-fusion algorithms.
The online version utilized all trajectory data, whereas the offline methods randomly sampled 300 points.

\paragraph{Baseline.}
The quasi-static experiments evaluated two categories of methods, consistent with the static analysis.
Reference-based methods estimate calibration parameters from fused orientation and accelerometer data,
including our ALAC (random mode), the same TLS method as in the static analysis, and online
implementations~\cite{al2023_imu_hand,troni2019_field}.
The angular-rate-aided method~\cite{troni2019_field} estimates only sensor bias ${^{s}\!A_0}$, while the Kalman
filter-based approach~\cite{al2023_imu_hand} assumes small-angle ${^{b}\!A_g}$ and excludes ${^{e}_{s}\!R}$ alignment rotation.
Self-calibration methods perform offline estimation of sensor parameters via iterative optimization,
including Newton's, LM, TR-Exact, and non-iterative ellipsoid fitting.
All self-calibration baselines were implemented consistently with the static experiments
(Table~\ref{tab_static_comparison}) using the same cost function and optimization setup.

\paragraph{Analysis.}
ALAC method clearly outperforms other reference-based approaches and online calibration
methods, and, under the average-filter metric, achieves accuracy comparable to the offline
self-calibration results, as illustrated in Table~\ref{tab_imu_comparison}. Under the average-filter
condition, the self-calibration methods (Newton's, LM, TR-Exact, and Ellipsoid) yield the lowest
RMSEs, consistent with the static experiments, as they are only affected by accelerometer uncertainty.
Under noisy raw measurements, our estimators achieves
the best RMSE and $\epsilon_\text{comp}$ among all baselines, while the iterative methods reach
local minima even initialized by our estimate.
Notably, our online implementation is slightly better than the offline one for both filtered and raw data.
These results demonstrate the robustness and reliability of our approach against the system uncertainties.

\section{Conclusion} \label{section_conclusion}
This work presents \textbf{ALAC}, a closed-form linear calibration algorithm for triaxial MEMS
accelerometers. It provides a unified attitude-aided framework that avoids the conventional strict
alignment and nonlinear or iterative calibration methods.
The calibration is formulated as a constrained homogeneous least-squares (CHLS) problem under
static gravity and efficiently solved using standard linear algebraic techniques such as GEVP and SVD.
A combined error matrix (CEM) compactly represents intrinsic sensor
errors, from which the scale factors, non-orthogonality, and alignment rotation are analytically derived.
The bias and gravity vector are estimated jointly, yielding a complete, self-consistent calibration model
applicable to embedded and robotic inertial platforms with attitude information.

Experiments demonstrate that ALAC maintains competitive accuracy under both attitude and
acceleration uncertainties. On a robot-mounted accelerometer, it outperforms conventional
reference-based calibration methods and achieves high precision with 24 selected orientations
comparable to self-calibration, demonstrating robustness to limited attitude accuracy.
On quasi-static IMU trajectories, both offline and online implementations achieve accuracy
higher than reference-based and comparable to best self-calibration on filtered data,
and outperform all baselines on raw measurements.
The recursive ALAC exhibits superior real-time stability and convergence in dynamic conditions,
enabling robust factory and in-field IMU accelerometer calibration for inertial navigation and sensor fusion
applications. The results indicate strong robustness against sensor, attitude, and mechanical uncertainties.

\paragraph{Practical implications.}
The proposed approach provides a practical and efficient alternative to traditional laboratory
calibration methods that rely on high-precision equipment, achieving higher accuracy by leveraging
the six-position orientations. Its recursive formulation also
outperforms the online baseline algorithms. For other offline cases, self-calibration via ellipsoid fitting
suffices; then separately calibrates the alignment between the accelerometer and platform frame
if needed. Their closed-form design make it suitable for the factory, laboratory and in-field calibration.

\begin{table}[t]
    \centering
    \small
    \caption{
        Dynamic-trajectory calibration errors for ALAC and baselines.
        Newton's/LM/TR-Exact initialized by our estimate;
        40-sample average filter (RMSE 0.2207 vs raw 0.3307);
        $\epsilon_\text{est}$ relative to Ours (reference).
        $^{\dagger}$TLS excludes ${^{e}_{s}\!R}$;
        Kalman-filter-based~\cite{al2023_imu_hand} assumes small-angle ${^{b}\!A_g}$ and excludes ${^{e}_{s}\!R}$;
        Angular-rate-aided~\cite{troni2019_field} estimates only ${^{s}\!A_0}$.
    }
    \label{tab_imu_comparison}
    \setlength{\tabcolsep}{3.8pt}
    \renewcommand{\arraystretch}{1.1}
    \begin{tabular}{llccccccccc}
        \toprule
        & & \multicolumn{5}{c}{Reference-based ($T$, $S$, ${^s\!A_0}$, ${^b\!A_g}$, ${^e_sR}$)$^{\dagger}$} & \multicolumn{4}{c}{Self-calibration ($T$, $S$, ${^s\!A_0}$)} \\
        \cmidrule(lr){3-7} \cmidrule(lr){8-11}
        & & \textbf{Ours} & TLS~\cite{duchi2023_tls_cuboid,duchi2024_imu_tls} & \textbf{Ours} (Online) & Online~\cite{al2023_imu_hand} & Online~\cite{troni2019_field} & Newton's & LM & TR-Exact & \textbf{Ellipsoid} \\
        \midrule
        \multirow{3}{*}{\shortstack{raw\\data}}
        & $\epsilon_\text{rmse}$ & 0.2755 & 0.3809 & 0.2751 & 0.3285 & 1.3654 & 0.0000 & 0.0000 & 0.0000 & 1.062 \\
        & $\epsilon_\text{comp}$ & 0.3965 & 0.4891 & 0.3915 & 0.5980 & - & - & - & - & - \\
        & $\epsilon_\text{est}$ & ref. & 0.381 & 0.206 & 0.824 & 1.800 & 66.195 & 6.29e6 & 36.641 & 4.874 \\
        \noalign{\vskip 1pt} \cdashline{1-11}[0.9pt/2pt] \noalign{\vskip 1pt}
        \multirow{3}{*}{\shortstack{avg.\\filter\\data}}
        & $\epsilon_\text{rmse}$ & 0.0996 & 0.1877 & 0.0962 & 0.2176 & 0.2953 & 0.0771 & 0.0771 & 0.0771 & 0.0770 \\
        & $\epsilon_\text{comp}$ & 0.207 & 0.326 & 0.206 & 0.485 & - & - & - & - & - \\
        & $\epsilon_\text{est}$ & ref. & 0.301 & 0.086 & 0.725 & 0.343 & 0.252 & 0.252 & 0.252 & 0.248 \\
        \noalign{\vskip 1pt} \cdashline{1-11}[0.9pt/2pt] \noalign{\vskip 1pt}
        \multirow{3}{*}{both}
        & time (s) & 5e-3 & 2e-3 & 7e-2 & 4e-1 & 1e-1 & 9e-2 & 8e-3 & 2e-2 & 2e-4 \\
        & data n & 300 & 300 & 5044 & 5044 & 5044 & 300 & 300 & 300 & 300 \\
        & sensitive & - & - & - & yes & - & yes & yes & yes & - \\
        \bottomrule
    \end{tabular}
\end{table}

\section*{Acknowledgment}
The algorithm presented in this paper was originally proposed in the Master's degree thesis
completed by Yongqiang Yu at the Harbin Institute of Technology in 2020.

\bibliographystyle{unsrtnat}
\bibliography{references}

@article{harindranath2024_survey,
  title = {A systematic review of user - conducted calibration methods for MEMS-based IMUs},
  journal = {Measurement},
  volume = {225},
  pages = {114001},
  year = {2024},
  doi = {10.1016/j.measurement.2023.114001},
  author = {Aparna Harindranath and Manish Arora}
}

@Article{ru2022mems_survey,
  AUTHOR = {Ru, Xu and Gu, Nian and Shang, Hang and Zhang, Heng},
  TITLE = {MEMS Inertial Sensor Calibration Technology: Current Status and Future Trends},
  JOURNAL = {Micromachines},
  VOLUME = {13},
  YEAR = {2022},
  NUMBER = {6},
  ARTICLE-NUMBER = {879},
  PubMedID = {35744491},
  DOI = {10.3390/mi13060879}
}

@article{poddar2016_survey,
  author = {Poddar, Shashi and Kumar, Vipan and Kumar, Amod},
  title = {A Comprehensive Overview of Inertial Sensor Calibration Techniques},
  journal = {Journal of Dynamic Systems, Measurement, and Control},
  volume = {139},
  number = {1},
  pages = {011006},
  year = {2016},
  month = {09},
  doi = {10.1115/1.4034419}
}

@article{David2007,
  title = {Calibration and data fusion solution for the miniature attitude and heading reference system},
  journal = {Sensors and Actuators A: Physical},
  volume = {138},
  number = {2},
  pages = {411-420},
  year = {2007},
  doi = {10.1016/j.sna.2007.05.008},
  author = {David Jurman and Marko Jankovec and Roman Kamnik and Marko Topi\v{c}}
}

@book{GolubVanLoan2013,
  author    = {Gene H. Golub and Charles F. Van Loan},
  title     = {Matrix Computations},
  publisher = {Johns Hopkins University Press},
  edition   = {4th},
  year      = {2013},
  address   = {Baltimore, MD},
  isbn      = {978-1421407944},
  doi       = {10.56021/9781421407944}
}

@book{BoydVandenberghe2004,
  author    = {Stephen Boyd and Lieven Vandenberghe},
  title     = {Convex Optimization},
  publisher = {Cambridge University Press},
  year      = {2004},
  address   = {Cambridge, UK},
  isbn      = {978-0521833783},
  url       = {https://web.stanford.edu/~boyd/cvxbook/}
}

@article{rohac2015_newton,
  title={Calibration of low-cost triaxial inertial sensors},
  author={Rohac, Jan and Sipos, Martin and Simanek, Jiri},
  journal={IEEE Instrumentation \& Measurement Magazine},
  volume={18},
  number={6},
  pages={32--38},
  year={2015},
  doi={10.1109/MIM.2015.7335836}
}

@ARTICLE{soriano2020,
  author={Soriano, Mario A. and Khan, Faheem and Ahmad, Rafiq},
  journal={IEEE Transactions on Instrumentation and Measurement},
  title={Two-Axis Accelerometer Calibration and Nonlinear Correction Using Neural Networks: Design, Optimization, and Experimental Evaluation},
  year={2020},
  volume={69},
  number={9},
  pages={6787-6794},
  doi={10.1109/TIM.2020.2978568}
}

@INPROCEEDINGS{kim2004initial,
  author={Kim, A. and Golnaraghi, M.F.},
  booktitle={PLANS 2004. Position Location and Navigation Symposium (IEEE Cat. No.04CH37556)},
  title={Initial calibration of an inertial measurement unit using an optical position tracking system},
  year={2004},
  volume={},
  number={},
  pages={96-101},
  doi={10.1109/PLANS.2004.1308980}
}

@inproceedings{wang2023mems_turntable,
  title={MEMS-IMU automatic calibration system design and implementation},
  author={Wang, Jingxiao and Liu, Ning},
  booktitle={Journal of Physics: Conference Series},
  volume={2492},
  number={1},
  pages={012005},
  year={2023},
  doi={10.1088/1742-6596/2492/1/012005}
}

@ARTICLE{won2010_valid_init,
  author={Won, Seong-hoon Peter and Golnaraghi, Farid},
  journal={IEEE Transactions on Instrumentation and Measurement},
  title={A Triaxial Accelerometer Calibration Method Using a Mathematical Model},
  year={2010},
  volume={59},
  number={8},
  pages={2144-2153},
  doi={10.1109/TIM.2009.2031849}
}

@article{cai2013,
  doi = {10.1088/0957-0233/24/10/105002},
  year = {2013},
  month = {aug},
  publisher = {IOP Publishing},
  volume = {24},
  number = {10},
  pages = {105002},
  author = {Cai, Qingzhong and Song, Ningfang and Yang, Gongliu and Liu, Yiliang},
  title = {Accelerometer calibration with nonlinear scale factor based on multi-position observation},
  journal = {Measurement Science and Technology}
}

@article{poddar2019_pso,
  title = {Scale-free PSO for in-run and infield inertial sensor calibration},
  journal = {Measurement},
  volume = {147},
  pages = {106849},
  year = {2019},
  doi = {10.1016/j.measurement.2019.07.077},
  author = {Shashi Poddar and Amod Kumar}
}

@INPROCEEDINGS{pesti2023_pso,
  author={Pesti, Richard and Sarcevic, Peter and Cs\'{i}k, Dominik and Odry, \'{A}kos},
  booktitle={2023 IEEE 17th International Symposium on Applied Computational Intelligence and Informatics (SACI)},
  title={Particle swarm optimization aided calibration of sensor installation errors for MEMS accelerometers},
  year={2023},
  volume={},
  number={},
  pages={493-498},
  doi={10.1109/SACI58269.2023.10158655}
}

@INPROCEEDINGS{cui2017_ga,
  author={Cui, Xiaole and Liu, Chunliang and Shi, Guangyi and Jin, Yufeng},
  booktitle={2017 IEEE International Conference on Real-time Computing and Robotics (RCAR)},
  title={A new calibration method for MEMS accelerometers with genetic algorithm},
  year={2017},
  volume={},
  number={},
  pages={240-245},
  doi={10.1109/RCAR.2017.8311867}
}

@INPROCEEDINGS{pesti2025_ga_robot,
  author={Pesti, Richard and Sarcevic, Peter and Cs\'{i}k, Dominik and Takacs, Marta and Odry, \'{A}kos},
  booktitle={2025 IEEE 19th International Symposium on Applied Computational Intelligence and Informatics (SACI)},
  title={Adaptive Neuro Fuzzy Inference System-Based Error Compensation for Mems Accelerometers},
  year={2025},
  volume={},
  number={},
  pages={1-6},
  doi={10.1109/SACI66288.2025.11030088}
}

@ARTICLE{sipos2012,
  author={Sipos, Martin and Paces, Pavel and Rohac, Jan and Novacek, Petr},
  journal={IEEE Sensors Journal},
  title={Analyses of Triaxial Accelerometer Calibration Algorithms},
  year={2012},
  volume={12},
  number={5},
  pages={1157-1165},
  doi={10.1109/JSEN.2011.2167319}
}

@INPROCEEDINGS{tedaldi2014_lm,
  author={Tedaldi, David and Pretto, Alberto and Menegatti, Emanuele},
  booktitle={2014 IEEE International Conference on Robotics and Automation (ICRA)},
  title={A robust and easy to implement method for IMU calibration without external equipments},
  year={2014},
  volume={},
  number={},
  pages={3042-3049},
  doi={10.1109/ICRA.2014.6907297}
}

@ARTICLE{frosio2012_lm,
  author={Frosio, Iuri and Pedersini, Federico and Borghese, N. Alberto},
  journal={IEEE Sensors Journal},
  title={Autocalibration of Triaxial MEMS Accelerometers With Automatic Sensor Model Selection},
  year={2012},
  volume={12},
  number={6},
  pages={2100-2108},
  doi={10.1109/JSEN.2012.2182991}
}

@INPROCEEDINGS{lu2016_gradient_descent,
  author={Lu, Xin and Liu, Zhong and He, Jingbo},
  booktitle={2016 8th International Conference on Intelligent Human-Machine Systems and Cybernetics (IHMSC)},
  title={Maximum Likelihood Approach for Low-Cost MEMS Triaxial Accelerometer Calibration},
  year={2016},
  volume={01},
  number={},
  pages={179-182},
  doi={10.1109/IHMSC.2016.184}
}

@ARTICLE{zou2024_gnls,
  author={Zou, Zuhao and Li, Liang and Hu, Xiangcheng and Zhu, Yilong and Xue, Bohuan and Wu, Jin and Liu, Ming},
  journal={IEEE Transactions on Instrumentation and Measurement},
  title={Robust Equipment-Free Calibration of Low-Cost Inertial Measurement Units},
  year={2024},
  volume={73},
  number={},
  pages={1-12},
  doi={10.1109/TIM.2023.3234081}
}

@article{botero2017lowcost,
  title={A low-cost platform based on a robotic arm for parameters estimation of Inertial Measurement Units},
  author={Botero-Valencia, Juan and Marquez-Viloria, David and Castano-Londono, Luis and Morantes-Guzmán, Luis},
  journal={Measurement},
  volume={110},
  pages={257--262},
  year={2017},
  doi = {10.1016/j.measurement.2017.07.002}
}

@ARTICLE{ye2017_linearization,
  author={Ye, Lin and Guo, Ying and Su, Steven W.},
  journal={IEEE Transactions on Instrumentation and Measurement},
  title={An Efficient Autocalibration Method for Triaxial Accelerometer},
  year={2017},
  volume={66},
  number={9},
  pages={2380-2390},
  doi={10.1109/TIM.2017.2706479}
}

@article{gietzelt201362_ellipsoid,
  title = {Performance comparison of accelerometer calibration algorithms based on 3D-ellipsoid fitting methods},
  journal = {Computer Methods and Programs in Biomedicine},
  volume = {111},
  number = {1},
  pages = {62-71},
  year = {2013},
  doi = {10.1016/j.cmpb.2013.03.006},
  author = {Matthias Gietzelt and Klaus-Hendrik Wolf and Michael Marschollek and Reinhold Haux}
}

@book{titterton2004_turntable,
  author = {David Titterton  and John Weston},
  title = {Strapdown Inertial Navigation Technology},
  publisher = {The Institution of Engineering and Technology},
  year = {2004},
  doi = {10.1049/PBRA017E},
  edition = {2nd}
}

@article{syed2007_turntable,
  doi = {10.1088/0957-0233/18/7/016},
  year = {2007},
  month = {may},
  volume = {18},
  number = {7},
  pages = {1897},
  author = {Syed, Z F and Aggarwal, P and Goodall, C and Niu, X and El-Sheimy, N},
  title = {A new multi-position calibration method for MEMS inertial navigation systems},
  journal = {Measurement Science and Technology}
}

@article{fang2014_turntable,
  author={Bin Fang and Wusheng Chou and Li Ding},
  title={An Optimal Calibration Method for a MEMS Inertial Measurement Unit},
  journal={International Journal of Advanced Robotic Systems},
  volume={11},
  number={2},
  pages={14},
  year={2014},
  doi={10.5772/57516}
}

@Article{luczak2022_turntable,
  AUTHOR = {\L{}uczak, Sergiusz and Zams, Maciej and D\k{a}browski, Bogdan and Kusznierewicz, Zbigniew},
  TITLE = {Tilt Sensor with Recalibration Feature Based on MEMS Accelerometer},
  JOURNAL = {Sensors},
  VOLUME = {22},
  YEAR = {2022},
  NUMBER = {4},
  ARTICLE-NUMBER = {1504},
  PubMedID = {35214402},
  DOI = {10.3390/s22041504}
}

@article{grip2011_linear_bs,
  doi = {10.1088/0957-0233/22/12/125103},
  year = {2011},
  month = {oct},
  publisher = {},
  volume = {22},
  number = {12},
  pages = {125103},
  author = {Grip, Niklas and Sabourova, Natalia},
  title = {Simple non-iterative calibration for triaxial accelerometers},
  journal = {Measurement Science and Technology}
}

@Article{cina2019_gnss_meams,
  AUTHOR = {Cina, Alberto and Manzino, Ambrogio Maria and Bendea, Iosif Horea},
  TITLE = {Improving GNSS Landslide Monitoring with the Use of Low-Cost MEMS Accelerometers},
  JOURNAL = {Applied Sciences},
  VOLUME = {9},
  YEAR = {2019},
  NUMBER = {23},
  ARTICLE-NUMBER = {5075},
  DOI = {10.3390/app9235075}
}

@ARTICLE{Renk2005,
  author={Renk, E.L. and Rizzo, M. and Collins, W. and Lee, F. and Bernstein, D.S.},
  journal={IEEE Control Systems Magazine},
  title={Calibrating a triaxial accelerometer-magnetometer - using robotic actuation for sensor reorientation during data collection},
  year={2005},
  volume={25},
  number={6},
  pages={86-95},
  doi={10.1109/MCS.2005.1550155}
}

@ARTICLE{beravs2012,
  author={Beravs, Tadej and Podobnik, Janez and Munih, Marko},
  journal={IEEE Transactions on Instrumentation and Measurement},
  title={Three-Axial Accelerometer Calibration Using Kalman Filter Covariance Matrix for Online Estimation of Optimal Sensor Orientation},
  year={2012},
  volume={61},
  number={9},
  pages={2501-2511},
  doi={10.1109/TIM.2012.2187360}
}

@article{khaula2025_robot,
  author       = {Hakim, Khaula Nurul and Nugroho, Yuniarto Wimbo and Rahardiyanti, Kandi and Rahmat, Mirza Zulfikar and Wicaksono, Bagus},
  title        = {Leveraging A Robotic Arm Platform for Low-Cost Calibration of Inertial Sensors on LAPAN Sounding Rockets},
  journal      = {International Journal on Advanced Science, Engineering and Information Technology},
  volume       = {15},
  number       = {2},
  pages        = {456--463},
  year         = {2025},
  month        = {April},
  doi          = {10.18517/ijaseit.15.2.20593}
}

@article{dong2020_cuboid,
  title={Calibration of low cost IMU's inertial sensors for improved attitude estimation},
  author={Dong, Mingjie and Yao, Guodong and Li, Jianfeng and Zhang, Leiyu},
  journal={Journal of Intelligent \& Robotic Systems},
  volume={100},
  number={3},
  pages={1015--1029},
  year={2020},
  doi={10.1007/s10846-020-01259-0}
}

@ARTICLE{xu2021novel,
  author={Xu, Tongxu and Xu, Xiang and Xu, Dacheng and Zhao, Heming},
  journal={IEEE Transactions on Instrumentation and Measurement},
  title={A Novel Calibration Method Using Six Positions for MEMS Triaxial Accelerometer},
  year={2021},
  volume={70},
  number={},
  pages={1-11},
  doi={10.1109/TIM.2020.3026024}
}

@InProceedings{duchi2023_tls_cuboid,
  author={Duchi, Massimo and Zaccaria, Federico and Briot, S{\'e}bastien and Ida', Edoardo},
  editor={Okada, Masafumi},
  title={Total Least Squares In-Field Identification for MEMS-Based Triaxial Accelerometers},
  booktitle={Advances in Mechanism and Machine Science},
  year={2023},
  publisher={Springer Nature Switzerland},
  address={Cham},
  pages={570--579},
  doi={10.1007/978-3-031-45770-8_57}
}

@INPROCEEDINGS{lv2016_nl_tls_cuboid,
  author={Lv, Jixin and Ravankar, Ankit A. and Kobayashi, Yukinori and Emaru, Takanori},
  booktitle={2016 IEEE/SICE International Symposium on System Integration (SII)},
  title={A method of low-cost IMU calibration and alignment},
  year={2016},
  volume={},
  number={},
  pages={373-378},
  doi={10.1109/SII.2016.7844027}
}

@ARTICLE{belkhouche2019_2accel,
  author={Belkhouche, Fethi},
  journal={IEEE Transactions on Instrumentation and Measurement},
  title={A Differential Accelerometer System: Offline Calibration and State Estimation},
  year={2019},
  volume={68},
  number={9},
  pages={3109-3118},
  doi={10.1109/TIM.2018.2876776}
}

@ARTICLE{lu2022_fog_imu,
  author={Lu, Jiazhen and Ye, Lili and Zhang, Jingxian and Luo, Wei and Liu, Haiqiao},
  journal={IEEE Sensors Journal},
  title={A New Calibration Method of MEMS IMU Plus FOG IMU},
  year={2022},
  volume={22},
  number={9},
  pages={8728-8737},
  doi={10.1109/JSEN.2022.3160692}
}

@ARTICLE{troni2019_field,
  author={Troni, Giancarlo and Whitcomb, Louis L.},
  journal={IEEE/ASME Transactions on Mechatronics},
  title={Field Sensor Bias Calibration With Angular-Rate Sensors: Theory and Experimental Evaluation With Application to Magnetometer Calibration},
  year={2019},
  volume={24},
  number={4},
  pages={1698-1710},
  doi={10.1109/TMECH.2019.2920367}
}

@ARTICLE{qureshi2017_newton,
  author={Qureshi, Umar and Golnaraghi, Farid},
  journal={IEEE Sensors Journal},
  title={An Algorithm for the In-Field Calibration of a MEMS IMU},
  year={2017},
  volume={17},
  number={22},
  pages={7479-7486},
  doi={10.1109/JSEN.2017.2751572}
}

@Article{li2019_newton,
  AUTHOR = {Li, Sen and Niu, Yunchen and Feng, Chunyong and Liu, Haiqiang and Zhang, Dan and Qin, Hengjie},
  TITLE = {An Onsite Calibration Method for MEMS-IMU in Building Mapping Fields},
  JOURNAL = {Sensors},
  VOLUME = {19},
  YEAR = {2019},
  NUMBER = {19},
  ARTICLE-NUMBER = {4150},
  PubMedID = {31557838},
  DOI = {10.3390/s19194150}
}

@ARTICLE{durr2023_bayesian,
  author={Dürr, Oliver and Fan, Po-Yu and Yin, Zong-Xian},
  journal={IEEE Sensors Journal},
  title={Bayesian Calibration of MEMS Accelerometers},
  year={2023},
  volume={23},
  number={12},
  pages={13319-13326},
  doi={10.1109/JSEN.2023.3272907}
}

@ARTICLE{al2023_imu_hand,
  author={Al Jlailaty, Hussein and Celik, Abdulkadir and Mansour, Mohammad M. and Eltawil, Ahmed M.},
  journal={IEEE Transactions on Instrumentation and Measurement},
  title={IMU Hand Calibration for Low-Cost MEMS Inertial Sensors},
  year={2023},
  volume={72},
  number={},
  pages={1-16},
  doi={10.1109/TIM.2023.3301860}
}

@article{chao2022_ellipsoid,
  doi = {10.1088/1361-6501/ac3ec2},
  year = {2021},
  month = {dec},
  publisher = {IOP Publishing},
  volume = {33},
  number = {2},
  pages = {025103},
  author = {Chao, Cui and Zhao, Jiankang and Zhu, Jianbin and Bessaad, Nassim},
  title = {Minimum settings calibration method for low-cost tri-axial IMU and magnetometer},
  journal = {Measurement Science and Technology}
}

@Article{duchi2024_imu_tls,
  AUTHOR = {Duchi, Massimo and Ida', Edoardo},
  TITLE = {Total Least Squares In-Field Identification for MEMS-Based Inertial Measurement Units},
  JOURNAL = {Robotics},
  VOLUME = {13},
  YEAR = {2024},
  NUMBER = {11},
  ARTICLE-NUMBER = {156},
  DOI = {10.3390/robotics13110156}
}

@misc{imu_dataset,
  author       = {Miguel Rasteiro},
  title        = {IMU Dataset},
  howpublished = {\url{https://github.com/miguelrasteiro/IMU_dataset}},
  note         = {Accessed: 2025-10-03},
  year         = {2019}
}

@article{Ahmad2013IMUReview,
  author    = {Norhafizan Ahmad and Raja Ariffin Raja Ghazilla and Nazirah M. Khairi and Vijayabaskar Kasi},
  title     = {Reviews on Various Inertial Measurement Unit (IMU) Sensor Applications},
  journal   = {International Journal of Signal Processing Systems},
  volume    = {1},
  number    = {2},
  pages     = {256--262},
  year      = {2013},
  month     = {December},
  doi       = {10.12720/ijsps.1.2.256-262}
}

@article{Johnston2019,
  author    = {William Johnston and Martin O'Reilly and Rob Argent and Brian Caulfield},
  title     = {Reliability, Validity and Utility of Inertial Sensor Systems for Postural Control Assessment in Sport Science and Medicine Applications: A Systematic Review},
  journal   = {Sports Medicine},
  year      = {2019},
  volume    = {49},
  number    = {5},
  pages     = {783--818},
  doi       = {10.1007/s40279-019-01095-9}
}

@Article{zhao2020_satellite,
  AUTHOR = {Zhao, Wanliang and Cheng, Yuxiang and Zhao, Sihan and Hu, Xiaomao and Rong, Yijie and Duan, Jie and Chen, Jiawei},
  TITLE = {Navigation Grade MEMS IMU for A Satellite},
  JOURNAL = {Micromachines},
  VOLUME = {12},
  YEAR = {2021},
  NUMBER = {2},
  ARTICLE-NUMBER = {151},
  PubMedID = {33557116},
  DOI = {10.3390/mi12020151}
}

@Article{mi12111373,
  AUTHOR = {Liu, Mei and Cai, Yuanli and Zhang, Lihao and Wang, Yiqun},
  TITLE = {Attitude Estimation Algorithm of Portable Mobile Robot Based on Complementary Filter},
  JOURNAL = {Micromachines},
  VOLUME = {12},
  YEAR = {2021},
  NUMBER = {11},
  ARTICLE-NUMBER = {1373},
  PubMedID = {34832785},
  DOI = {10.3390/mi12111373}
}

@article{ZHANG2019454,
  title = {Low-cost IMU and odometer tightly coupled integration with Robust Kalman filter for underground 3-D pipeline mapping},
  journal = {Measurement},
  volume = {137},
  pages = {454-463},
  year = {2019},
  doi = {10.1016/j.measurement.2019.01.068},
  author = {Penghe Zhang and Craig Matthew Hancock and Lawrence Lau and Gethin Wyn Roberts and Huib {de Ligt}},
}

@article{WANG2017111,
  title = {Augmented Cubature Kalman filter for nonlinear RTK/MIMU integrated navigation with non-additive noise},
  journal = {Measurement},
  volume = {97},
  pages = {111-125},
  year = {2017},
  doi = {10.1016/j.measurement.2016.10.056},
  author = {Dingjie Wang and Hanfeng Lv and Jie Wu},
}

\appendix

\section{Derivation of the GEVP} \label{app_gevp}
Consider the quadratic constrained quadratic program
\begin{equation}
    \min_{x}\;\|\bar{A} x\|_2^2 \quad \text{s.t. } \|Dx\|_2^2=1 \quad\rightarrow\quad
    \min_{x}\; x^\top G x \quad \text{s.t. } x^\top B x = 1 ,
\end{equation}
with $G=\bar{A}^\top \bar{A}\succeq 0$ and $B=D^\top D\succeq 0$.
Applying the method of Lagrange multipliers,
\begin{equation}
    \mathcal{L}(x,\lambda) = x^\top G x - \lambda (x^\top B x - 1)
\end{equation}
and the stationarity condition
\begin{equation}
    \nabla_x \mathcal{L} = 2Gx - 2\lambda Bx = 0
\end{equation}
leads directly to the generalized eigenvalue problem (GEVP)
\begin{equation}
    Gx = \lambda Bx .
\end{equation}
The optimal solution $x^*$ is the eigenvector corresponding to the smallest finite generalized
eigenvalue, normalized such that $x^{*\top}Bx^*=1$. Existing numerical solvers (e.g.,
\texttt{scipy.linalg.eig} or MATLAB \texttt{eig}) can be directly applied.

\section{Recursive GEVP Formulation} \label{app_recursive}
Building on the generalized eigenvalue formulation in Appendix~\ref{app_gevp},
at time step $n$, the calibration problem is
\begin{equation}
    \min_x\; x^\top G_n x
    \quad \text{s.t. } x^\top Bx = 1,
\end{equation}
where $\bar{A}_n\!\in\!\mathbb{R}^{3n\times15}$ stacks all measurements up to step $n$,
$G_n=\bar{A}_n^\top\bar{A}_n$, and $B=D^\top D$.

When a new measurement block $A_{n+1}\!\in\!\mathbb{R}^{3\times15}$ arrives,
the Gram matrix is updated recursively as
\begin{equation}
    \label{equ_recursive_gevp}
    G_{n+1} = G_n + A_{n+1}^\top A_{n+1}.
\end{equation}
This avoids reconstructing $\bar{A}_{n+1}$ explicitly and requires only a rank-15 update per step.
The problem at step $n\!+\!1$ remains a GEVP with the updated $G_{n+1}$,
and its smallest eigenpair provides the refined calibration estimate.

\section{Equivalent SEVP} \label{app_sevp}
Starting from \eqref{equ_J}, assuming $x=\begin{bmatrix}x_{12}; x_3\end{bmatrix}$,
a new variable $y=\tfrac{1}{g}I_{3}\cdot x_3=P\cdot x_3$ is defined as
\begin{equation}
    \label{equ_x3}
    x_3=P^{-1}\cdot y .
\end{equation}
Substituting \eqref{equ_x3} into \eqref{equ_AX_definision} gives
\begin{equation}
    \bar{A}x=[\bar{A}_{12},\bar{A}_3]
    \begin{bmatrix}
    x_{12} \\ x_3
    \end{bmatrix}
    =\bar{A}_{12}x_{12}+\bar{A}_3x_3
    =\bar{A}_{12}x_{12}+\bar{A}_3P^{-1}\cdot y ,
\end{equation}
where $\bar{A}_{12}\in\mathbb{R}^{3n\times 12}$ and $\bar{A}_3\in\mathbb{R}^{3n\times 3}$
denote the block partitions of $\bar{A}$.
Therefore, \eqref{equ_J} is equivalently written as
\begin{equation}
    \label{equ_J_y}
    \min_{x_{12},y} \; J = \|\bar{A}_{12}x_{12}+\bar{A}_3P^{-1}\cdot y \|_2^2
    \quad \text{s.t. } \|y\|_2 = 1 .
\end{equation}
For fixed $y$, the optimal $x_{12}$ is obtained from the normal equations:
\begin{equation}
    \label{equ_x12}
    x_{12}=-(\bar{A}_{12}^\top \bar{A}_{12})^{-1}\bar{A}_{12}^\top \bar{A}_3P^{-1}\cdot y .
\end{equation}
Substituting \eqref{equ_x12} into \eqref{equ_J_y} yields a standard eigenvalue problem (SEVP)
with respect to $y$:
\begin{equation}
    \label{equ_Ay}
    \min_{y} \; J = \|Hy\|_2^2
    \quad \text{s.t. } \|y\|_2 = 1 ,
\end{equation}
where $
H = \big(\bar{A}_3 - \bar{A}_{12}(\bar{A}_{12}^\top \bar{A}_{12})^{-1}\bar{A}_{12}^\top \bar{A}_3\big)P^{-1}
$.
The optimal solution of \eqref{equ_Ay} is $y^*$, the eigenvector corresponding to
the minimum eigenvalue of $H^\top H$ \cite{GolubVanLoan2013,BoydVandenberghe2004}.

Finally, the optimal solution $x^*$ of \eqref{equ_J} is recovered from \eqref{equ_x12} and
\eqref{equ_x3} as
\begin{equation}
    \label{equ_x_solution}
    x^* =
    \begin{bmatrix}
    -(\bar{A}_{12}^\top \bar{A}_{12})^{-1}\bar{A}_{12}^\top \bar{A}_3P^{-1}\cdot y^* \\
    P^{-1}\cdot y^*
    \end{bmatrix}_{15\times 1} .
\end{equation}

\section{Orientations for the six-position strategy} \label{app_24_orientations}
The following 24 orientations $^b_eR_i$ $(i=1,\ldots,24)$ are divided into four groups.

\paragraph{Group 1.} (1-2 fix col-1 to $\pm[0,0,1]^\top$; 3-4 fix col-2; 5-6 fix col-3)
\[
\begin{bmatrix}
    0 & 0 & -1 \\
    0 & +1 & 0 \\
    +1 & 0 & 0
\end{bmatrix},\;
\begin{bmatrix}
    0 & 0 & -1 \\
    0 & -1 & 0 \\
    -1 & 0 & 0
\end{bmatrix},\;
\begin{bmatrix}
    +1 & 0 & 0 \\
    0 & 0 & -1 \\
    0 & +1 & 0
\end{bmatrix},\;
\begin{bmatrix}
    +1 & 0 & 0 \\
    0 & 0 & +1 \\
    0 & -1 & 0
\end{bmatrix},\;
\begin{bmatrix}
    -1 & 0 & 0 \\
    0 & -1 & 0 \\
    0 & 0 & +1
\end{bmatrix},\;
\begin{bmatrix}
    -1 & 0 & 0 \\
    0 & +1 & 0 \\
    0 & 0 & -1
\end{bmatrix}
\]

\paragraph{Group 2.} (1-2 fix col-1 to $\pm[0,0,1]^\top$; 3-4 fix col-2; 5-6 fix col-3)
\[
\begin{bmatrix}
0 & +1 & 0 \\
0 & 0 & +1 \\
+1 & 0 & 0
\end{bmatrix},\;
\begin{bmatrix}
0 & +1 & 0 \\
0 & 0 & -1 \\
-1 & 0 & 0
\end{bmatrix},\;
\begin{bmatrix}
0 & 0 & +1 \\
+1 & 0 & 0 \\
0 & +1 & 0
\end{bmatrix},\;
\begin{bmatrix}
0 & 0 & +1 \\
-1 & 0 & 0 \\
0 & -1 & 0
\end{bmatrix},\;
\begin{bmatrix}
0 & -1 & 0 \\
+1 & 0 & 0 \\
0 & 0 & +1
\end{bmatrix},\;
\begin{bmatrix}
0 & -1 & 0 \\
-1 & 0 & 0 \\
0 & 0 & -1
\end{bmatrix}
\]

\paragraph{Group 3.} (1-2 fix col-1 to $\pm[0,0,1]^\top$; 3-4 fix col-2; 5-6 fix col-3)
\[
\begin{bmatrix}
0 & 0 & +1 \\
0 & -1 & 0 \\
+1 & 0 & 0
\end{bmatrix},\;
\begin{bmatrix}
0 & 0 & +1 \\
0 & +1 & 0 \\
-1 & 0 & 0
\end{bmatrix},\;
\begin{bmatrix}
-1 & 0 & 0 \\
0 & 0 & +1 \\
0 & +1 & 0
\end{bmatrix},\;
\begin{bmatrix}
-1 & 0 & 0 \\
0 & 0 & -1 \\
0 & -1 & 0
\end{bmatrix},\;
\begin{bmatrix}
+1 & 0 & 0 \\
0 & +1 & 0 \\
0 & 0 & +1
\end{bmatrix},\;
\begin{bmatrix}
+1 & 0 & 0 \\
0 & -1 & 0 \\
0 & 0 & -1
\end{bmatrix}
\]

\paragraph{Group 4.} (1-2 fix col-1 to $\pm[0,0,1]^\top$; 3-4 fix col-2; 5-6 fix col-3)
\[
\begin{bmatrix}
0 & -1 & 0 \\
0 & 0 & -1 \\
+1 & 0 & 0
\end{bmatrix},\;
\begin{bmatrix}
0 & -1 & 0 \\
0 & 0 & +1 \\
-1 & 0 & 0
\end{bmatrix},\;
\begin{bmatrix}
0 & 0 & -1 \\
-1 & 0 & 0 \\
0 & +1 & 0
\end{bmatrix},\;
\begin{bmatrix}
0 & 0 & -1 \\
+1 & 0 & 0 \\
0 & -1 & 0
\end{bmatrix},\;
\begin{bmatrix}
0 & +1 & 0 \\
-1 & 0 & 0 \\
0 & 0 & +1
\end{bmatrix},\;
\begin{bmatrix}
0 & +1 & 0 \\
+1 & 0 & 0 \\
0 & 0 & -1
\end{bmatrix}
\]


\end{document}